\documentclass[conf]{new-aiaa}
\usepackage[utf8]{inputenc}
\DeclareCaptionFont{xbf}{\bfseries\boldmath}
\usepackage{graphicx, caption, subcaption}
\usepackage{amsmath}
\usepackage[version=4]{mhchem}
\usepackage{siunitx}
\usepackage{longtable,tabularx}

\title{A Multi-fidelity Double-Delta Wing Dataset and Empirical Scaling Laws for GNN-based Aerodynamic Field Surrogates}

\author{Yiren Shen\footnote{Ph.D. Candidate, Department of Aeronautics and Astronautics, AIAA Student Member.} and Juan J. Alonso\footnote{Vance D. and Arlene C. Coffman Professor, Department of Aeronautics and Astronautics, AIAA Fellow.}}
\affil{Stanford University, Stanford, CA 94305, USA} 

\footnotetext{Submitted to AIAA SciTech Forum 2026.}

\begin{document}

\maketitle

\begin{abstract}

Data-driven surrogate models are increasingly adopted to accelerate vehicle design. However, open-source multi-fidelity datasets and empirical guidelines linking dataset size to model performance remain limited. This study investigates the relationship between training data size and prediction accuracy for a graph neural network (GNN) based surrogate model for aerodynamic field prediction. We release an open-source, multi-fidelity aerodynamic dataset for double-delta wings, comprising 2448 flow snapshots across 272 geometries evaluated at angles of attack from $11^\circ$ to $19^\circ$ at $Ma=0.3$ using both \acf{VLM} and \acf{RANS} solvers. The geometries are generated using a nested Saltelli sampling scheme to support future dataset expansion and variance-based sensitivity analysis. Using this dataset, we conduct a preliminary empirical scaling study of the \texttt{MF-VortexNet} surrogate by constructing six training datasets with sizes ranging from 40 to 1280 snapshots and training models with 0.1 to 2.4 million parameters under a fixed training budget. We find that the test error decreases with data size with a power-law exponent of $\mathbf{-0.6122}$, indicating efficient data utilization. Based on this scaling law, we estimate that the optimal sampling density is approximately eight samples per dimension in a d-dimensional design space. The results also suggest improved data utilization efficiency for larger surrogate models, implying a potential trade-off between dataset generation cost and model training budget.

\end{abstract}

\begin{acronym}
    \acro{CFD}{computational fluid dynamics}
\acro{FVM}{finite-volume method}
\acro{HF}{high-fidelity}
\acro{LF}{low-fidelity}
\acro{MDO}{multidisciplinary design optimization}
\acro{MF}{multi-fidelity}
\acro{ML}{machine learning}
\acro{MSE}{mean-square error}
\acro{PDE}{partial differential equation}
\acro{RANS}{Reynolds-Averaged Navier-Stokes}
\acro{SCALOS}{Supersonic Configurations at Low Speeds}
\acro{NASA}{National Aeronautics and Space Administration}
\acro{HSCT}{High Speed Supersonic Transport}
\acro{SST}{supersonic transport}
\acrodef{SC}[S\&C]{stability and control}
\acro{AOA}{angle-of-attack}
\acro{SA}{Spalart-Allmaras}
\acro{VLM}{vortex lattice method}
\acro{RNMP}{Reference Noise Measurement Points}
\acro{MAC}{mean aerodynamic chord}
\acro{QOI}{quantity of interests}
\acro{DOE}{design of experiments}
\acro{OML}{outer mold line}
\acro{GNN}{graph neural network}
\acro{NPI}{new-product introduction}
\acro{MSE}{mean squared error}


\end{acronym}

\section*{Nomenclature}

{\renewcommand\arraystretch{1.0}

\noindent\begin{longtable*}{l @{\quad=\quad} p{6.3cm}l }
$AR$ & wing aspect ratio $AR = S/b^2$ \\
$B$ & wing span (inches)\\
$BW$ & inboard-outboard break chord location \\
$C_D$ & drag coefficient            \\     
$C_L$ & lift coefficient              \\   
$C_M$ & pitching moment coefficient     \\ 
$\Lambda_{in}$ & inboard leading-edge sweep angle ($^\circ$) \\
$\Lambda_{out}$ & outboard leading-edge sweep angle ($^\circ$) \\
$\Lambda_{te}$ & outboard trailing-edge sweep angle ($^\circ$) \\

\end{longtable*}

}
\section{Introduction}
\label{sec:introduction}

\lettrine{S}{urrogate} models are widely used in aircraft design and \ac{MDO} due to the complex nature of underlying physics and engineering problems \cite{Wissink-sage-2025}. As a result of the complexity, the assessments and optimizations of vehicles are costly when high-dimensional models or experiments are used as the means of performance assessment. The core idea of surrogate modeling is to construct a mathematically simplified function that captures a predefined set of \ac{QOI}, such as lift and drag coefficients $C_L$ and $C_D$, allowing the surrogate to replace costly high order evaluations \cite{willcox_uq}. By leveraging data-driven approximations, surrogate models enable rapid predictions while preserving essential physical trends, thereby significantly reducing the cost of design and analysis and accelerating the \ac{NPI} process. 

Recently, due to the advances in computational hardware and novel data-driven models, such as neural networks, a wide variety of new surrogate model methods has been proposed for aerospace applications \cite{Wassing-pinnfield-2025, Abras-dnnqoi-2025, Healy-dnnpod-2025, anand-polarpred-2025, kilic-svmdnn-2025, Inthra-mfgpr-2025, Lee-gpr-2025, Shen2025VortexNet, ZHAO2022105643}. These surrogate modeling techniques range from Gaussian Process Regression (GPR), Polynomial Chaos Expansion (PCE), Proper Orthogonal Decomposition (POD) and reduced order modeling (ROM), Deep Neural Networks (DNNs), to Physics Informed Neural Networks (PINNs). Each class of method is suited to different trade-offs among prediction accuracy, applicability, interpretability, and computational efficiency. Some methods, such as GPR and PCE, are more suitable for the prediction of a set of scalar quantities, while others, such as POD, PINNs, and DNNs, have the potential for predicting distributed field quantities such as pressure or velocity fields. 

The capability of offering complex mappings from design variables and flow conditions to spatially resolved fields is often desired in \ac{MDO} cycles. These field quantities provide richer design feedback for designers, allowing the coupling of analyses across disciplines and fidelities with multiple solvers, and enabling postprocessing of arbitrary \acp{QOI}. Popular techniques for field surrogate modeling include POD-Kriging \cite{Mrosek2019PODkriging} and convolutional neural network (CNN) \cite{Shen2024}. These methods often require a consistent spatial topology among surrogate data inputs; however, when the geometry undergoes large shape changes, maintaining such correspondence becomes difficult, thereby limiting their generalizability across the design space in \ac{MDO} applications.

To overcome this limitation, recent work has shifted toward unstructured geometry representations that operate directly on unstructured point clouds and graphs. Point-cloud-based models naturally encode both geometric and solution information as nodal or edge features, eliminating the need for mesh mapping. For example, \texttt{PointNet} architecture directly predicts surface pressure distributions from point clouds \cite{shen-pointnet-2023}, while graph neural network (GNN) surrogates such as \texttt{MeshGraphNets} demonstrate accurate field reconstruction across widely varying geometries \cite{Pfaff2021MeshGraphNets}. In parallel, Implicit Neural Representations have been used for flow-field reconstruction \cite{Catalani2024}, while operator-learning frameworks such as DeepONet and DoMINO \cite{Lu_2021, ranade2025dominodecomposablemultiscaleiterative}, Geometry-Informed Neural Operators \cite{li2023geometryinformedneuraloperatorlargescale}, and Fourier Neural Operators (FNO) \cite{li2021fourierneuraloperatorparametric} have been applied to aerodynamics prediction and parametric flow modeling \cite{zhao-deeponet-2023, li2023geometryinformedneuraloperatorlargescale, li2021fourierneuraloperatorparametric}, providing mesh-independent mappings between function spaces. More recently, transformer-based surrogates have also been proposed for vehicle aerodynamics modeling \cite{alkin2025abuptscalingneuralcfd}. Collectively, these developments signal a broader transition from mesh-dependent field surrogates toward geometry- and discretization-agnostic surrogate models for aerodynamic design.

Beyond geometric generalization, it is also desirable that surrogate models support mappings across fidelity hierarchies. In \ac{NPI} process, it is common for analyses to be performed at different fidelities: from the initial conceptual design to detailed design, increasingly sophisticated higher-fidelity models are deployed for assessment and optimization. It is therefore desirable to have a mean of bridging the fidelity gaps across tools. Among the field-prediction approaches, graph-based physics-informed surrogates have shown particular promise for multi-fidelity mappings as these models are relatively inexpensive to train and the graph-based contextualization of data sources fits mesh-based analyses naturally. Previously, we proposed \texttt{MF-VortexNet}, a physics-informed multi-fidelity GNN surrogate that learns to map low-fidelity pressure fields computed by \ac{VLM} to high-fidelity pressure fields computed by \ac{CFD} on the same lifting-surface lattice \cite{Shen2025VortexNet}. When applied to low-speed, high \ac{AOA} Delta wing configurations featuring nonlinear fluid dynamics that are not modeled by the \ac{VLM} solver, \texttt{MF-VortexNet} successfully maps low-fidelity pressure loading to its high-fidelity counterpart and thus significantly reduces normalized errors in aerodynamic coefficients prediction, while generalizing reasonably across unseen geometries within a prescribed design space \cite{Shen2025SCALOS}. When integrated with a conceptual design environment, the model showed sufficient fidelity gain to enable stability-constrained aerodynamic shape optimization for a double-delta (cranked delta) wing test case \cite{shen_2025_aviationGNNASO}. 

However, questions remain regarding how \texttt{MF-VortexNet}'s prediction accuracy scales with training dataset size and what data requirements are necessary to reach a target accuracy level. In particular, empirical scaling laws \cite{kaplan2020scalinglawsneurallanguage} for physics-informed, multi-fidelity \ac{GNN} surrogates models for field prediction remain largely absent from the aerodynamic literature \cite{liu2024neuralscalinglawsgraphs}, hindering informed decisions about \ac{DOE} strategies and the application of similar tools in \ac{MDO}. 

To study how surrogate-model accuracy scales with dataset size, the underlying training dataset must contain both substantial geometric variability and sufficiently complex flow physics. Although the AIAA Applied Aerodynamics Surrogate Modeling (AASM) discussion group has recently released several benchmark datasets for surrogate modeling\cite{bekemeyer_benchmark}, these datasets do not provide the controlled geometric parametrization desired for the present scaling analysis. To address this need, we adopt a low-speed double-delta wing as the test configuration. This geometry offers a six-dimensional design space that enables systematic shape variations. Furthermore, the external flow over a Delta wing at higher \acp{AOA} is highly nonlinear and dominated by vortex-driven phenomena \cite{Seraj2022hiAOA}, making it an informative and challenging setting for evaluating the predictive capacity of field-based surrogate models. Finally, the chosen geometry maintains a balance between engineering realism and data-preprocessing simplicity: the \ac{OML} surfaces require no complex cleanup or fuselage-wing intersection handling, yet still represent a three-dimensional aerodynamic problem.

This work addresses these two interrelated gaps. First, we present an open-source, multi-fidelity aerodynamic dataset for double-delta wings across 272 unique geometries and multiple \acp{AOA} ranging from $11^\circ$ to $19^\circ$ at low speed ($Ma = 0.3$). The \ac{DOE} employs a nested Saltelli sampling scheme to generate geometric configurations, enabling nested geometric variations and supporting future variance-based sensitivity analysis and uncertainty quantification. Aerodynamic coefficients, surface solutions, and volume solutions are provided in open formats, along with the \ac{CFD} configuration files, to maximize downstream usability of the dataset. Second, we perform an empirical scaling study of the \texttt{MF-VortexNet} model using nested subsets of the dataset. By selecting between 8 and 256 geometric configurations for training set construction, we quantify a power-law decay of prediction error with increasing sampling density in the design space.

The remainder of this paper is organized as follows. Section~\ref{sec:methods} outlines the relevant technical framework of the different analysis components, including geometry generation environment, low- and high-fidelity solvers, and the \texttt{MF-VortexNet} surrogate model. Section~\ref{sec:dataset} presents a detailed description of the double-delta wing aerodynamic dataset, including the design space, \ac{DOE}, simulation configurations, and simulation results. Section~\ref{sec:scalingtest} presents the setup and results of the scaling study. Finally, Section~\ref{sec:conclusion} discusses the implications of the current study and outlines future work.

\section{Methodology}
\label{sec:methods}

As this study involves modeling external flow over wings with multiple solvers at different fidelities, and utilizing a machine learning model \texttt{MF-VortexNet} to learn the mapping from a pressure--geometry--flow-condition state to a high-fidelity pressure field, this section provides a brief discussion of the modules used throughout the study.

\subsection{Multi-fidelity Geometry Representation}
\label{ssec:methods-designenv}

The \texttt{SUAVE} conceptual design environment \cite{Lukaczyk2015suave} is used to generate vehicles from geometry parameters, drive low-fidelity simulations, and create surface-mesh tessellations using its built-in \texttt{OpenVSP} \cite{OpenVSP} interface. Depending on the subsequent applications, the model representations of a given vehicle configuration may have different fidelities.

At the lowest fidelity, an instance of the \texttt{Vehicle} class in \texttt{SUAVE} is created, taking in the corresponding design variable vector $\mathbf{x}\in\mathbf{R}^n$, where $n$ is the dimension of the design space, and a prescribed process that constructs a vehicle from $\mathbf{x}$. Details of the computer program structure, including the \texttt{Vehicle} class definition, can be found in \citeauthor{Lukaczyk2015suave} \cite{Lukaczyk2015suave}. When $\mathbf{x}$ is changed, different instances are created to represent different vehicle configurations.

When different analyses are conducted on the vehicle instance, different types of instance representation are needed. When the \ac{VLM} (see Section~\ref{ssec:methods-lfsolver}) is utilized, a lattice grid of the vortex distribution is created based on the \texttt{Vehicle} instance by incorporating the encoded geometric parameters. Additional configuration settings, such as the number of panels or the distribution functions used for the lattice discretization, may change the vehicle-geometry representation for a given instance. The lattice grid and vortex distribution are used by the \ac{VLM} solver to perform aerodynamic analyses.

Likewise, for CFD simulations, a representation of the \ac{OML} of the vehicle is required by the solver. Under such analyses, we use \texttt{SUAVE}'s \texttt{OpenVSP} \cite{OpenVSP} interface to translate the \texttt{SUAVE} instance to an \ac{OML} boundary representation. Subsequently, \texttt{OpenVSP}'s \texttt{ComputeCFDMesh} function is used to generate a tessellated surface mesh. This mesh is then imported into \texttt{Pointwise}, together with a far-field boundary defined with a diameter of 40 \acp{MAC}, to define the fluid domain. The fluid domain is then further discretized to generate the volume mesh for CFD analyses.

\subsection{Low-fidelity Solver}
\label{ssec:methods-lfsolver}

The \ac{VLM} solver solves the potential-flow problem for the perturbation velocity potential, subject to a no-penetration boundary condition on the lifting surface. The lifting surface is discretized into a quadrilateral panel lattice. Horseshoe vortices are defined on each panel's quarter-chord as well as its side bounds, with semi-infinite trailing legs extending downstream. The no-penetration boundary condition is enforced at the collocation points located at the center of three-quarter-chord of each panel. This leads to a linear system for the vortex strength of each panel's horseshoe vortex, which can be solved. The resultant vortex strengths at each panel can then be summed for induced-velocity computation using the Biot-Savart law \cite{anderson_fundamentals_2010}. The pressure perturbation due to the induced velocity is computed using Bernoulli's theorem \cite{anderson_fundamentals_2010}. The specific \ac{VLM} used in this work is shipped with \texttt{SUAVE}, which is based on \citeauthor{Miranda1976VORLAX}'s \texttt{VORLAX} implementation, and is suitable for compressibility-corrected, inviscid, attached flows \cite{Miranda1976VORLAX}.

Due to the compressibility correction, the \ac{VLM} predictions for vehicles at high speed and low \ac{AOA} is relatively accurate. The computational cost, at a chordwise-by-spanwise lattice of $30 \times 32$, is negligible. In addition, the solver's convergence is robust across a wide design space, making it ideal for rapid design space exploration studies. However, due to the neglect of nonlinear flow effects such as vortex lift and separation, the prediction accuracy deteriorates for vehicles at low speed under high \acp{AOA}. As shown by \citeauthor{Shen2025SCALOS} in their previous study comparing the \acp{QOI} computed from \ac{VLM} and \ac{RANS} \ac{CFD}, \ac{VLM} results for relevant Delta wing configurations are relatively accurate up to \ac{AOA} of $10^\circ$ \cite{Shen2025SCALOS}.

\subsection{High-fidelity Solver}
\label{ssec:methods-hfsolver}

The \texttt{SU2} \ac{RANS} \ac{CFD} solver is used to compute aerodynamics \cite{Economon2016SU2}. The solver solves the \ac{RANS} equations, in which nonlinear flow effects are accounted for. For turbulence modeling, the Spalart-Allmaras turbulence model with rotation corrections (SA-R) \cite{Dacles-mariani-SARC} is used. The SA-R turbulence model adds a one-equation eddy-viscosity closure to the \ac{RANS} equations, such that the eddy viscosity is adjusted across different flow regions based on the dominance of rotational and strain effects.

For the present application, the selected solver and turbulence model achieve a balance between the desired accuracy and cost. The chosen solver captures viscous effects and vortex systems for flow that is attached or mildly separated, which constitutes the majority of the flow conditions considered in the current study. Several previous works have demonstrated the suitability of using the SA-R \ac{RANS} solver for external-flow modeling with strong vortex presence \cite{Dacles-mariani-SARC, Alauzet2024sar23}. For Delta wings and double-delta wings, several studies have highlighted the capability of SA model family to predict vortex strength and breakdown \cite{Widhalm-sarc}, to model aerodynamics at higher \ac{AOA} with accurate capture of vortex-vortex interactions \cite{Alauzet2024sar23}, as well as to capture vortex lift-induced effects on surface pressure distribution over wings \cite{Seraj2022hiAOA, Widhalm-sarc, di2024sarc, Schütte-sarc}.

The chosen SA-R \ac{RANS} formulation performs well for attached and mildly separated flows, but its accuracy degrades in regimes dominated by massive separation or unsteady fluid dynamics. For double-delta wings, these effects become significant near the critical \acp{AOA} at approximately $18^\circ$ for some geometries, such that vortex sheet oscillation or separated flow are seen. As a result, higher-fidelity approaches such as detached-eddy simulation would be required to achieve reliable predictions. However, the computational cost of these methods is prohibitive to the authors for the broad design-space sampling required in the present study. Consequently, they are not included in the current dataset but represent a potential avenue for future multi-fidelity dataset construction.

\subsection{Multi-fidelity Surrogate Model: \texttt{MF-VortexNet}}
\label{ssec:methods-vortexnet}

A surrogate model is deployed to map low-fidelity panel-wise pressure coefficients solved by the \ac{VLM} solver to the corresponding pressure field of the \ac{CFD} solver on the same lifting-surface lattice. The surrogate model is \texttt{MF-VortexNet} \cite{Shen2025VortexNet}, a light-weight physics-informed \ac{GNN} machine-learning model. To contextualize field data, the \ac{VLM} lattice is represented by a graph, where nodes aggregate local geometric descriptors, flow conditions, and low-fidelity pressure-coefficient features, while edges encode spatial distances among panels. Within this framework, \texttt{MF-VortexNet} employs a U-Net-like encoder-decoder architecture built from multi-head graph-attention convolutional blocks \cite{veličković2018} to construct latent representations that capture pressure loading induced by nonlinear vortical structures and other complex flow phenomena, and then decodes these representations to corrected nodal pressure fields that approximate CFD solutions while maintaining robust prediction accuracy across a design space that is typical in the conceptual-design phase.

The surrogate model is embedded into a multi-fidelity aerodynamic workflow as described by \citeauthor{Shen2025VortexNet} to construct paired low- and high-fidelity fields for offline training \cite{Shen2025VortexNet}. In online assessment, the baseline fields computed by the \ac{VLM} solver are subsequently corrected by \texttt{MF-VortexNet}, thereby yielding ``quasi-CFD'' pressure distributions. Subsequently, derived force and moment coefficients can be computed from the augmented pressure loading at a computational cost comparable to that of the underlying \ac{VLM} solver. Because the model operates on full fields rather than on a small set of scalar \acp{QOI}, arbitrary integrated metrics can be obtained by post-processing the corrected pressure distribution, which improves both flexibility and interpretability of early-stage design studies, and provides a means to couple with solvers in other disciplines. When applied to low-speed, high \ac{AOA} double-delta wing configurations, the \texttt{MF-VortexNet}-augmented workflow demonstrates substantial design-fidelity improvement in capturing moment and stability performance relative to relying on \ac{VLM} alone, while maintaining generalization across unseen geometries and freestream conditions within a prescribed design space \cite{Shen2025SCALOS, shen_2025_aviationGNNASO}.

\section{Dataset}
\label{sec:dataset}

This section details the double-delta wing aerodynamic dataset. We first define the parametric double-delta wing geometry and its six-dimensional design space. Next, we describe the \ac{DOE}, which utilizes a nested Saltelli sampling scheme to ensure scalable sampling. Finally, we present our numerical simulation framework and results, including the automated meshing pipeline and solver configurations for generating the paired \ac{VLM} and \ac{CFD} data.

\subsection{Geometry Definition}
\label{ssec:dataset-geomDef}

\begin{figure}[!ht]
\centering
\resizebox{0.8\textwidth}{!}{%
    \begin{circuitikz}
    \tikzstyle{every node}=[font=\Large]
    \draw [dashed] (1.5,7.75) -- (13.75,7.75);
    \draw [dashed] (2.75,10.5) -- (13.75,10.5);
    \draw [dashed] (1.5,13.25) -- (13.75,13.25);
    \draw [line width=1pt, short] (3.25,7.75) -- (8.75,10.5);
    \draw [line width=1pt, short] (8.75,10.5) -- (11,13.25);
    \draw [line width=1pt, short] (11,13.25) -- (13,13.25);
    \draw [line width=1pt, short] (13,13.25) -- (12.25,10.5);
    \draw [line width=1pt, short] (12.25,10.5) -- (12.25,7.75);
    \draw [line width=0.6pt, <->, >=Stealth] (3,8.5) .. controls (3.5,8.5) and (3.75,8.25) .. (3.75,8);
    \draw [<->, >=Stealth] (8.75,11.75) .. controls (9.25,12) and (9.5,11.75) .. (9.5,11.5);
    \draw [<->, >=Stealth] (12.25,12.5) .. controls (12.5,12.5) and (12.5,12.5) .. (12.75,12.25);
    \draw [<->, >=Stealth] (1.5,7.75) -- (1.5,13.25);
    \draw [<->, >=Stealth] (3,10.5) -- (3,7.75);
    \node [font=\sffamily\Large] at (4,9.00) {\text{$\Lambda_{in}$}};
    \node [font=\sffamily\Large] at (9.5,12.25) {\text{$\Lambda_{out}$}};
    \node [font=\sffamily\Large] at (13.5,12.25) {\text{$\Lambda_{te}$}};
    \node [font=\sffamily\Large] at (0.75,9.5) {\text{0.5B}};
    \node [font=\sffamily\Large] at (2.25,9.5) {\text{BW2}};
    \draw [ color={rgb,255:red,255; green,0; blue,0} , fill={rgb,255:red,255; green,0; blue,0}] (3.2,7.75) circle (0.15cm);
    \draw [dashed] (8.75,10.5) -- (8.75,13.25);
    \draw [dashed] (12.25,10.5) -- (12.25,13.25);
    \draw [dashed, line width=1.5pt] (5.95, 7.75) -- (9.8, 10.5);
    \draw [dashed, line width=1.5pt] (9.8, 10.5) -- (11.6, 13.25);
    \node [font=\sffamily\Large] at (10, 9.75) {\text{$\delta$}};
    \end{circuitikz}
}%
\caption{Geometry definition and its design variables.}
\label{fig:wing_platform}
\end{figure}
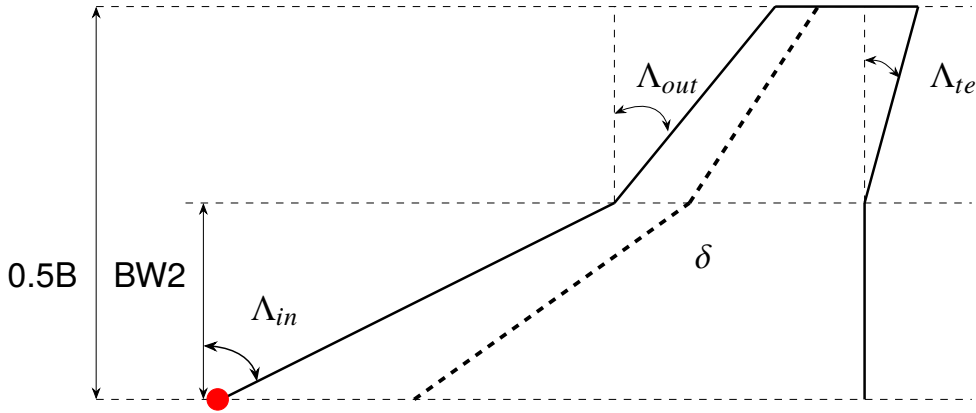

Six design variables are used to define the double-delta wing shape. Figure~\ref{fig:wing_platform} shows the half-wing about the central-body symmetry plane. The geometry has six design parameters, including B, the wing span (in inches); BW2, the fractional location of the inboard-to-outboard break-chord relative to the semi-span; SW1 ($\Lambda_{in}$), the inboard leading-edge sweep angle; SW2 ($\Lambda_{out}$), the outboard leading-edge sweep angle; SR2 ($\Lambda_{te}$), the outboard trailing-edge sweep angle; and DROOP ($\delta$), the camber-deflection angle applied to the baseline airfoil. The baseline airfoil is constructed from the root-airfoil geometry of the RW23 wing presented by \citeauthor{Wiersma_2025_cfd} \cite{Wiersma_2025_cfd} and is applied uniformly across the span.

To allow a richer representation of camber variation within the design space, $\delta$ acts by deflecting the baseline airfoil's leading edge camber about the $30\%$ chord from leading edge. The airfoil morphing process begins with cubic-spline fitting of the baseline airfoil's upper and lower surfaces from their discrete coordinates. The camber and thickness distributions are then extracted on a shared chordwise grid. The droop morphing is subsequently applied to the baseline camber, introducing $C^1$ and $C^2$ parametric discontinuities \cite{CS348reader} at the pivot point and producing two separate Bézier curves for the leading and trailing cambers, respectively. To ensure smoothness, the Bézier control-point parameters for the leading camber line are solved using a sequential least-squares programming (SLSQP) solver in \texttt{SciPy} to minimize the $C^1$ and $C^2$ continuity residuals between the leading- and trailing-camber curves at the pivot point. The modified camber is then combined with the original thickness distribution to reconstruct the upper and lower surfaces, with an additional round of smoothing applied to both surfaces' curves.

The spanwise twist distribution is defined as follows: the inboard section, from the root to the break-chord location, maintains a constant twist of $1.5^\circ$ leading edge up, while the outboard section applies a linear washout from $1.5^\circ$ to $-1.5^\circ$ toward the tip.

\vspace{15pt}

\subsection{Design Space}
\label{ssec:dataset-designSpace}

The design space $\mathbf{x}$ is defined as $\mathbf{x} = [\delta, \Lambda_{in}, \Lambda_{out}, \Lambda_{te}, BW2, B]^T, x_i \in [x_i^{min}, x_i^{max}]$, where $x_i^{min}$ and $x_i^{max}$ are the bounds. The selection of the design variable bounds is based on first referencing the design space of recent commercial supersonics technology program, including Supersonic Configurations at Low Speeds (SCALOS) \cite{nasa-scalos-report}, and High-Speed Civil Transport (HCST) \cite{Giunta_Balabanov_1997}. From these studies, the nominal inboard and outboard leading edge sweep ($\Lambda_{in}$ and $\Lambda_{out}$) is $68^\circ$ and $45^\circ$ respectively. The aspect ratio of the wing platform is taken to be $2.5$. From these nominal values, we consider the design space included in SCALOS program \cite{nasa-scalos-report}, and adjust the design variable bounds so that the created design space does not produce non-physical designs. Specifically, design variable combinations that lead to negative tip chord under our \ac{DOE}. This constraint reduces the design space represented in this dataset, resulting in a $\Lambda_{in}$ upper bound smaller than those explored in the SCALOS program. However, the constraint ensures that the resulting samples are balanced, as no non-physical designs are removed from the samples, in contrast to our previous experiments \cite{shen_2025_aviationGNNASO}. This design space is presented in Table ~\ref{tab:dv_bounds}. 

\begin{table}[h]
    \centering
    \begin{tabular}{lccc} \hline \hline
        \textbf{Design Variable} & \textbf{Lower} & \textbf{Upper} & \textbf{Nominal} \\\hline 
        $\delta$ ($^\circ$) & -8 & 7 & 0 \\
        $\Lambda_{in}$ ($^\circ$)   & 47 & 67 & 65 \\
        $\Lambda_{out}$ ($^\circ$)   & 35 & 60 & 45 \\
        $\Lambda_{te}$ ($^\circ$)   & 0 & 25 & 10 \\
        BW2    & 0.32 & 0.47 & 0.4 \\
        B (inches)     & 1100 & 1300 & 1200 \\
         \hline \hline
    \end{tabular}
    \caption{List of design variables and their corresponding upper and lower bounds.}
    \label{tab:dv_bounds}
\end{table}

\subsection{Design of Experiments}
\label{ssec:dataset-doe}

\subsubsection{Sampling Methods}

The \ac{DOE} involves the selection of combinations of design variables to generate geometry samples. We have three major goals for the \ac{DOE}. Firstly, we aim to achieve efficient and homogeneous design-space filling, such that the sampling scheme does not introduce sampling bias that leads to clustered design samples within the design space, which could bias the performance of subsequently trained surrogate models. Secondly, the sampling scheme should produce an extendable dataset, such that when the dataset needs to be expanded, new sampling points can be added without requiring existing experiments to be re-run. This property is particularly important when the experiment is costly, such as the \ac{CFD} simulations in this study. Finally, we want the sampling scheme to support future downstream assessments of the surrogate models trained from the dataset, including applications involving sensitivity analyses or uncertainty quantification.

Popular sampling methods are considered, including factorial sampling, random sampling, Latin Hypercube Sampling (LHS), and quasi-random sequences. The quasi-random sequence sampling methods, particularly Sobol sequence, meet the previously-mentioned \ac{DOE} goals better. Sobol sequence constructs a sequence of spatial quadrature points that enable fast convergence of n-dimensional integral, that is, to find a sequence  ${x_i}$ with length $P$ such that, 
\begin{equation}
    \frac{1}{P} \sum_{i=0}^{P-1} f(x_i),
\end{equation}
approximates the integration
\begin{equation}
    \int_{[0,1]^\text{n}} f(x) dx.
\end{equation}

Solving for the Sobol sequence directly is costly, and modern software packages use fast Sobol sequence algorithms in which subsequent Sobol numbers are generated by XOR-ing previous Sobol number with a predefined table of direction numbers \cite{stephensobolseuqence}. For the current \ac{DOE}, we adopt \texttt{SALib} \cite{salib2}, which contains a Sobol sequence generation routine by \citeauthor{stephensobolseuqence} \cite{stephensobolseuqence}. For a required $P$ samples in $n$ dimensional space, the Sobol sequence generator generates a base matrix $S$ of size $ (P + \text{skipvalue}) \times 2n$. The $\text{skipvalue}$ is applied to discard the first several rows of the Sobol sequence matrix, as repetitions may occur \cite{campolongo2011screening}, thereby improving the uniformity of samples generated. For a predefined skipvalue and problem dimension, the generated Sobol sequence is deterministic. Throughout the experiments, we set the skipvalue to 16.

From the base matrix, we follow Saltelli's sampling scheme \cite{saltelli1999quantitative} using \texttt{SALib}'s \texttt{sample.saltelli} function with function call of \texttt{saltelli.sample(problem, skipvalue=16, calc\_second\_order=False)}. The process firstly constructs two sampling matrices $A$ and $B$, such that
\begin{equation}
    A[i,:] \gets S[i, 0:n] \hspace{10pt}  \text{and} \hspace{10pt}
    B[i,:] \gets S[i, n:2n].
\end{equation}
Then assembling $n$ cross-sampling matrices $AB^{(k)}$ by cross swapping columns of $B$ to a copy of the $A$ matrix. Mathematically, this is equivalent to, 
\begin{equation}
    AB^{(k)} = (A[:,1],...,A[:,k-1], B[:,k], A[:,k+1], ... , A[:,n])  \hspace{5pt} \text{for} \hspace{5pt} k\in [1,n].
\end{equation}
The final design samples matrix $\tilde{X}$ in unit hypercube is constructed by concatenating all $A$, $B$, and $AB$ matrices vertically, such that, 
\begin{equation}
    \tilde{X} = \begin{bmatrix}
    A \\
    AB^{(1)} \\
    \vdots \\
    AB^{(n)} \\
    B
    \end{bmatrix}
    \in \mathbb{R}^{P(n+2) \times n} .
\end{equation}
The final design samples $X$ can be obtained from $\tilde{X}$ by column-wise scaling using the upper-and-lower bounds of the corresponding design variables. 

The adopted sampling scheme meets our \ac{DOE} goals. Firstly, the Saltelli sampling scheme is deterministic and reproducible. By generating samples from a Sobol sequence, the method achieves uniform space-filling properties that minimize clustering and sampling bias, leading to balanced datasets ideal for machine-learning model training. Secondly, because the Sobol sequence is nested, doubling the dataset size yields a refined design-space coverage without modifying the existing design points, ensuring that dataset scaling does not require rerunning existing experiments. Moreover, the Saltelli framework naturally supports variance-based sensitivity analysis and uncertainty quantification, thereby enriching the downstream applications of the dataset.

The Saltelli sequence samples $X$ are used exclusively for model training. Because Saltelli sampling enforces a homogeneous distribution, randomly partitioning $X$ to form a test set would disrupt this structure and introduce bias into the remaining training samples. To avoid this issue, we generate a separate holdout test set independent of $X$ with 16 geometries. The holdout test set is generated using LHS sampling via \texttt{SALib}'s \texttt{sample.latin} function \cite{salib2}. The resulting unit-hypercube samples are subsequently scaled using the same procedure employed in the Saltelli sampling process described previously.

\subsubsection{Sampling Results}

As a result of the sampling scheme described, one can generate geometry configurations at different levels of density in a nested structure. We took six values of $P$, in multiples of 2, to generate six levels of datasets, shown in Table~\ref{tab:dataset-levels}. Each geometry in the dataset is evaluated at nine \acp{AOA}, ranging from $11^\circ$ to $19^\circ$ in increments of $1^\circ$.

\begin{table}[ht]
    \centering
    \begin{tabular}{lcccc} \hline \hline
        \textbf{Level} & \textbf{Base Sequence Length ($P$)} & \textbf{Geometry Configurations}  & \textbf{Number of AOAs} & \textbf{Dataset Size ($D$)}\\\hline 
            1 & 1 & 8 & 9   &  72 \\
            2 & 2 & 16 & 9  & 144\\
            3 & 4 & 32 & 9  & 288\\
            4 & 8 & 64 & 9  & 576\\
            5 & 16 & 128 & 9& 1152\\
            6 & 32 & 256 & 9& 2304\\
         \hline \hline
    \end{tabular}
    \caption{Dataset sizes at six levels.}
    \label{tab:dataset-levels}
\end{table}

\begin{figure}[htbp!]
    \centering

    \begin{minipage}{0.8\textwidth}
        \centering
        \textbf{(a) Top View} \\[1ex]
        \includegraphics[width=\textwidth]{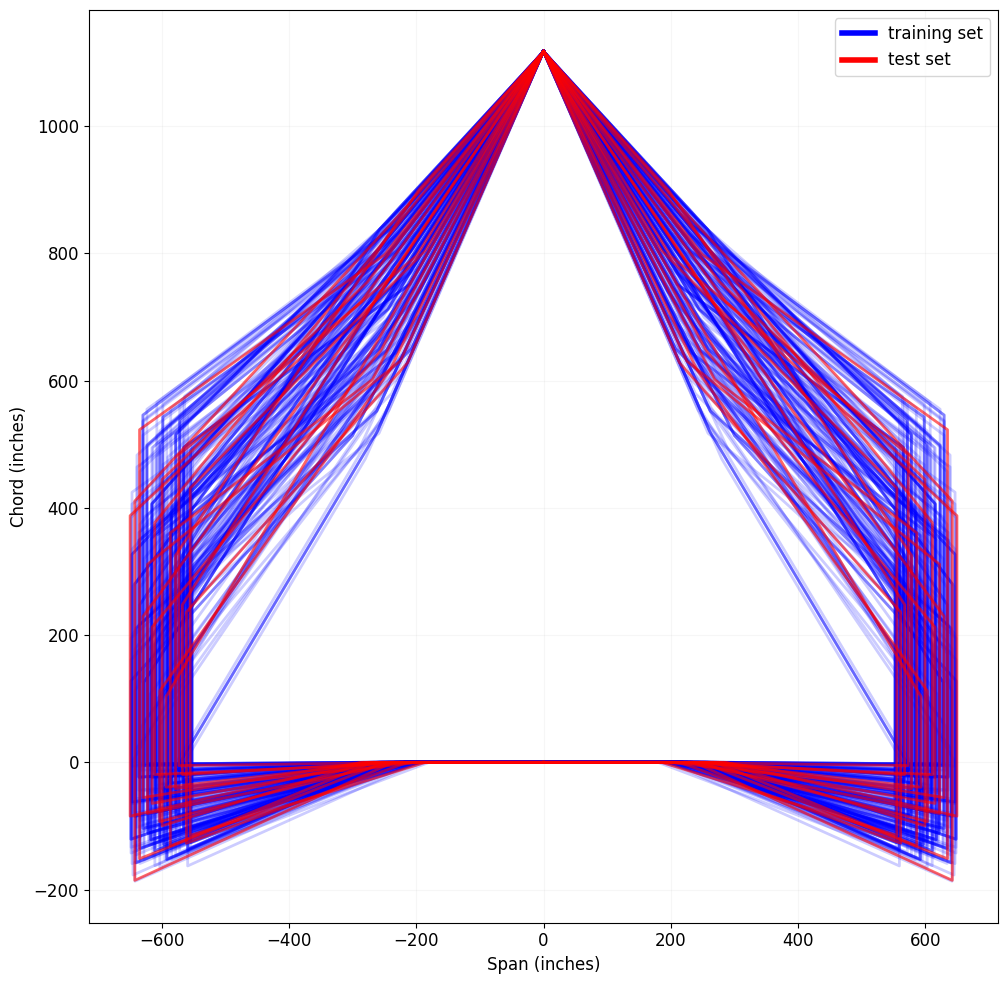}
    \end{minipage}

    \vspace{1.5em} 

    \begin{minipage}{0.9\textwidth}
        \centering
        \textbf{(b) Side View} \\[1ex]
        \includegraphics[width=\textwidth]{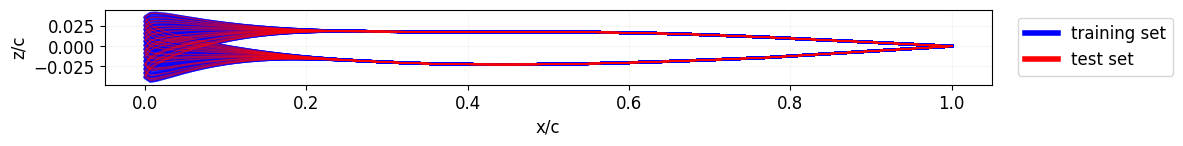}
    \end{minipage}

    \caption{Illustration of vehicle configurations sampled from the design space. Training samples from the Level~6 set are shown in blue, while samples from the holdout test set are shown in red. }
    \label{fig:dspace-drawing}
\end{figure}

Together, these configurations constitute of the training geometry set of the dataset. Due to the nested structure of the sampling scheme, the set of geometries in the Level~1 dataset is a strict subset of those in the Level~2 dataset, and this nested structure is preserved across all higher dataset levels. The largest dataset is designated as the Level~6 set. All designs in the Level~6 set are shown in Figure~\ref{fig:dspace-drawing}(a) in top view, and the corresponding side view at the root in Figure~\ref{fig:dspace-drawing} (b). In both figures, the blue edges denote the configurations included in the training geometry set, and the red edges indicate the configurations belonging to the holdout test set. 

From Figure~\ref{fig:dspace-drawing}(a), it is evident that the training set spans a wide design space, ranging from low-aspect-ratio trapezoidal wings (with the largest tip-chord lengths) to delta wings (with the smallest tip-chord lengths). The sampling density is relatively uniform across the design space. Similarly, the holdout test set also exhibits a consistent sampling density, as indicated by the even distribution of red edges across the design space. Similarly, uniform design space coverage for $\delta$ can be observed in Figure~\ref{fig:dspace-drawing}(b).

\begin{figure}[htbp!]
    \centering
    \includegraphics[width=0.96\linewidth, trim={0cm 0cm 0cm 3cm}, clip]{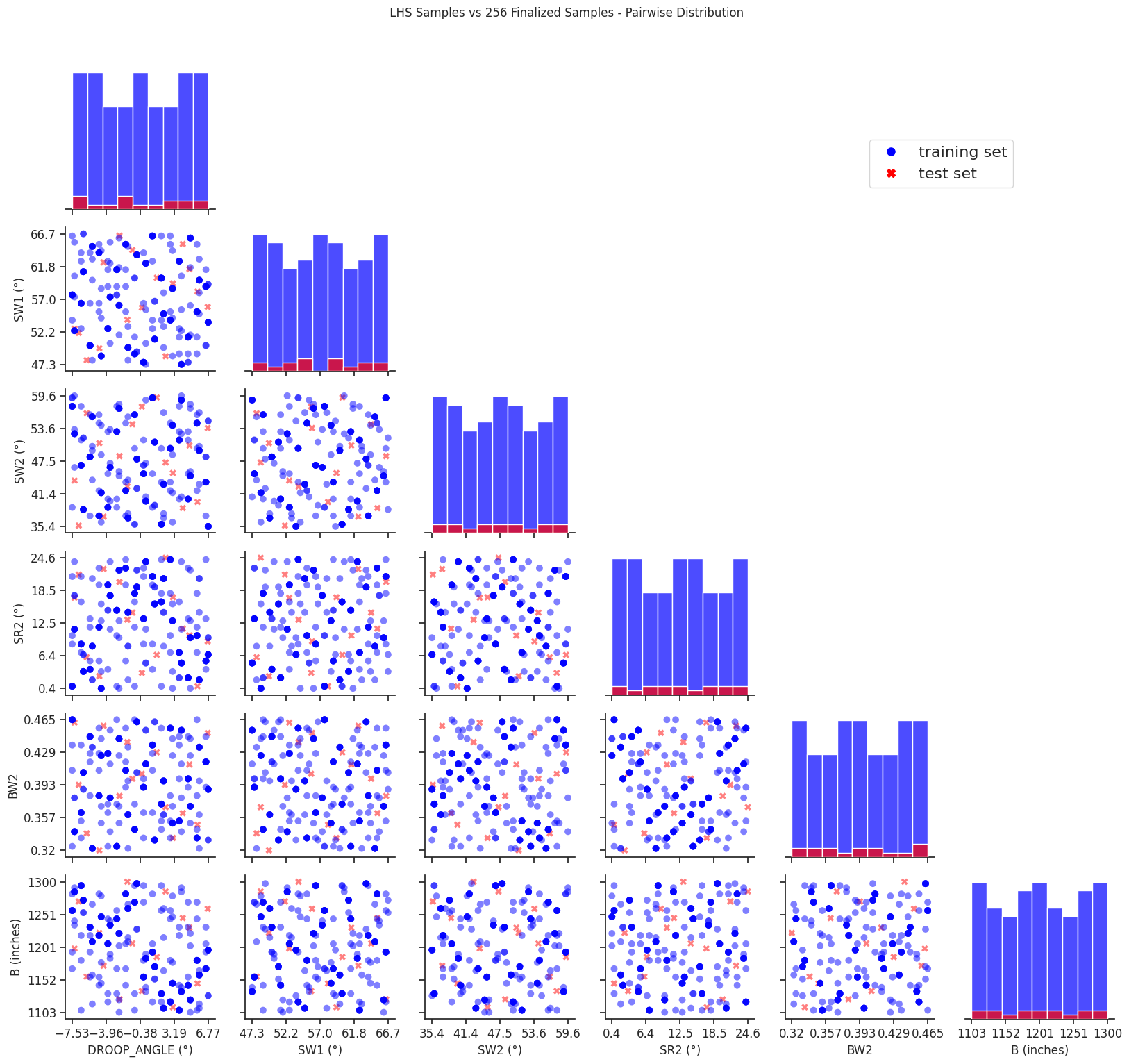}
    \caption{Design space coverage and design variable distributions of the dataset.}
    \label{fig:dspace-pairplot}
\end{figure}

To further illustrate the sampling density, Figure~\ref{fig:dspace-pairplot} presents the pair plots of the design variables for samples in the Level~6 set and holdout set. The off-diagonal panels show the scatter plots of paired design variables, while the diagonal panels show the histograms of the sampling density with respect to each design variable. Samples belonging to the training geometry set are plotted in blue, while samples corresponding to the holdout test set are shown in red. As indicated by the paired plots, no specific clustering is observed across any design variable pairs, and within each design variable, a uniform sampling density is achieved across the range of design-variable bounds. When examining the distribution of the holdout test set, the test samples also span the design space evenly. Because the sampling scheme for the training set is nested, the sampling density distribution across training set levels is consistent, and hence the sampling density pair plots for the sparser sampling levels are not shown.

\subsection{Numerical Simulation}
\label{ssec:dataset-simulation}

\subsubsection{CFD Meshing}
Vehicle geometries are generated from the design variable samples using \texttt{SUAVE}'s \texttt{write\_vsp\_mesh} interface to produce the corresponding \texttt{OpenVSP} geometry model, as discussed in Section~\ref{ssec:methods-designenv}. The \ac{CFD} mesh generation involves two stages: surface meshes are first produced using \texttt{OpenVSP}'s \texttt{ComputeCFDMesh} function, and volume meshes are then created in \texttt{Pointwise} using its \texttt{T-Rex} extrusion function. The \texttt{T-Rex} procedure extrudes the surface mesh into the freestream to form an anisotropic boundary layer mesh with controlled growth and resolution suitable for \ac{RANS} simulations.

\begin{figure}[h!]
    \begin{minipage}{0.33\textwidth}
        \centering
        \textbf{(a) Level 1} \\[1ex]
        \includegraphics[width=\textwidth, trim={20cm 0cm 20cm 0cm}, clip]{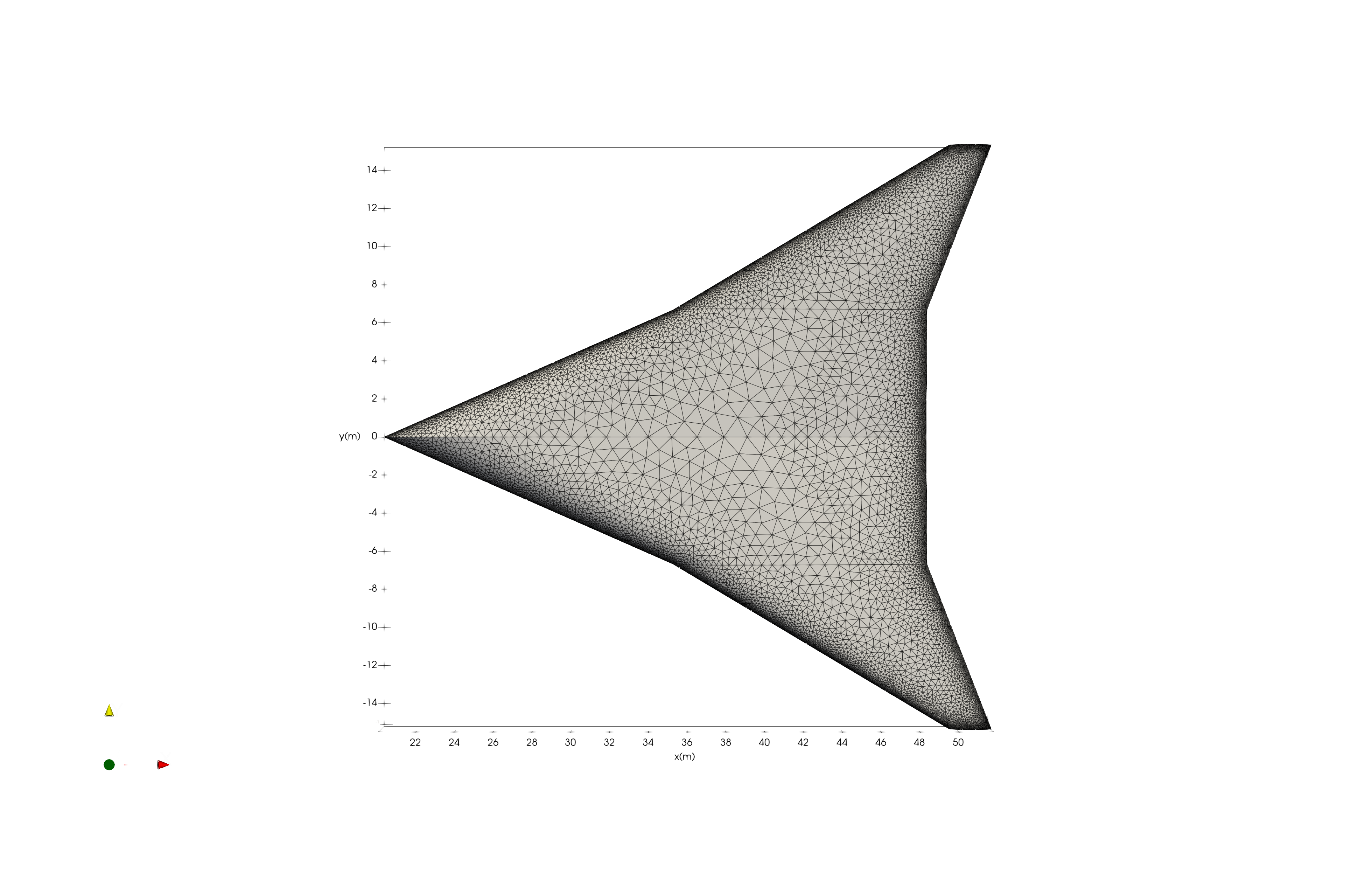}
    \end{minipage}
    \hfill
    \begin{minipage}{0.33\textwidth}
        \centering
        \textbf{(b) Level 2} \\[1ex]
        \includegraphics[width=\textwidth, trim={20cm 0cm 20cm 0cm}, clip]{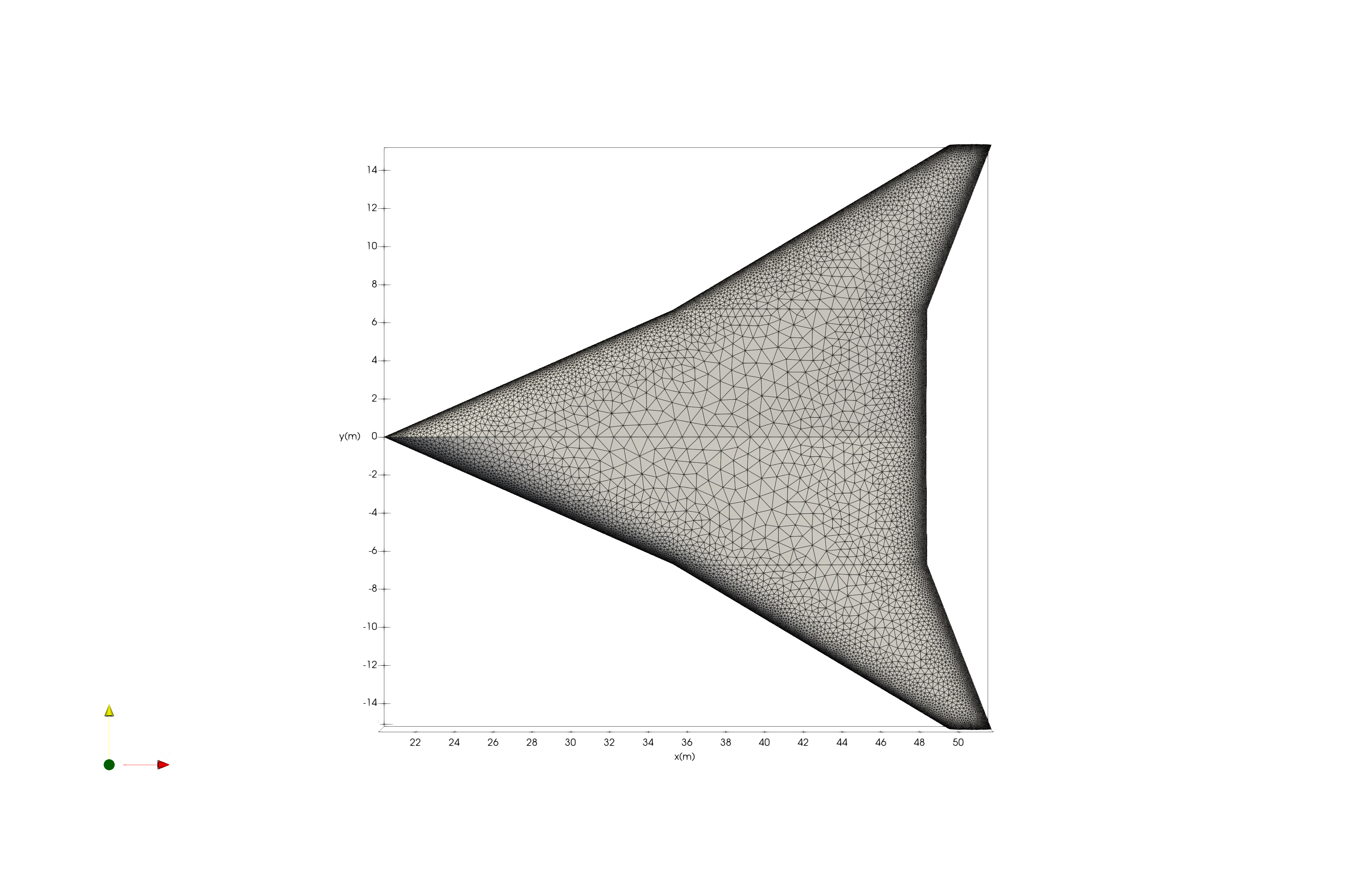}
    \end{minipage}
    \hfill
    \begin{minipage}{0.33\textwidth}
        \centering
        \textbf{(c) Level 3} \\[1ex]
        \includegraphics[width=\textwidth, trim={20cm 0cm 20cm 0cm}, clip]{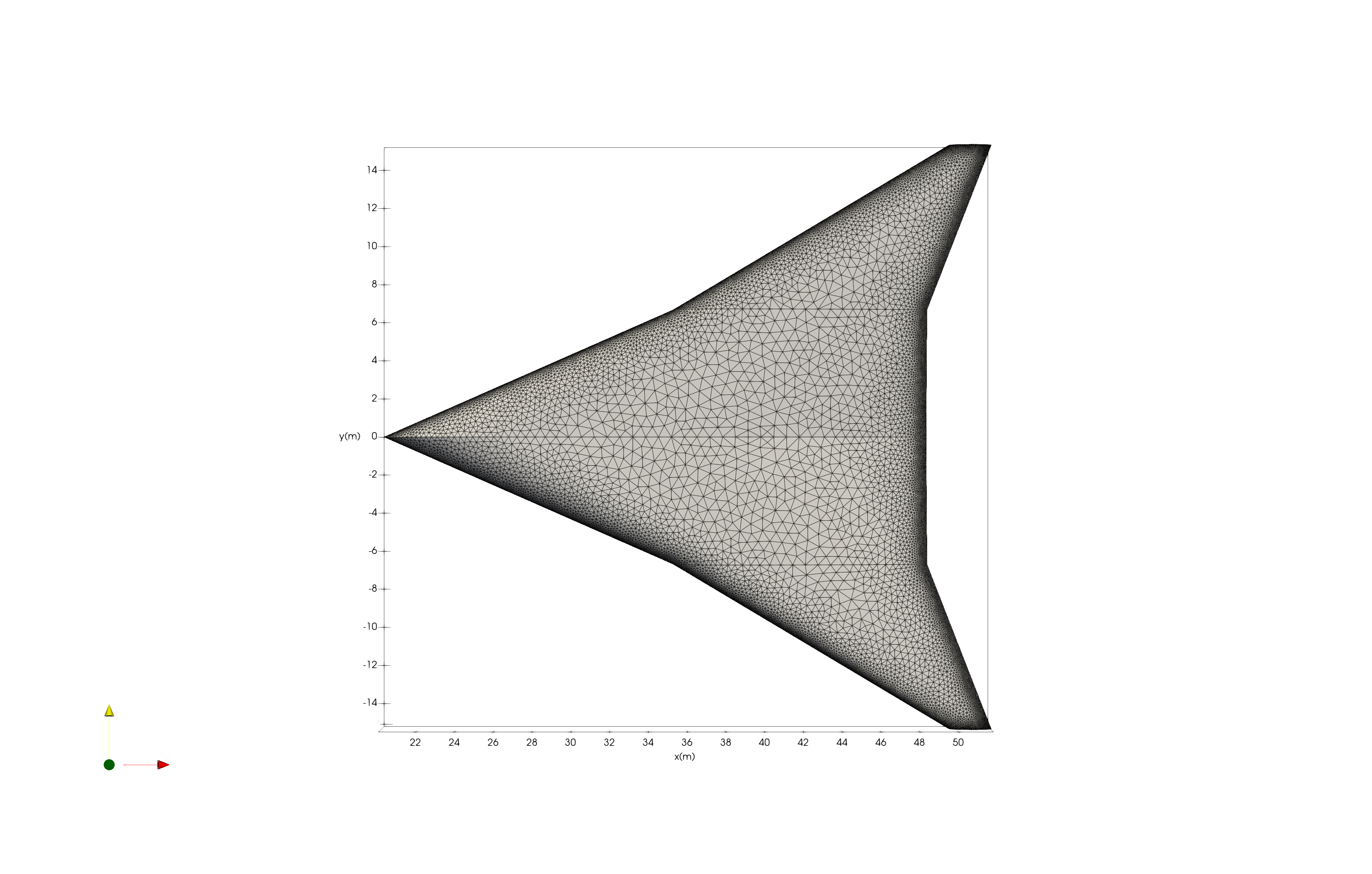}
    \end{minipage}
    \vspace{1pt}
    \begin{minipage}{0.33\textwidth}
        \centering
        \textbf{(d) Level 4} \\[1ex]
        \includegraphics[width=\textwidth, trim={20cm 0cm 20cm 0cm}, clip]{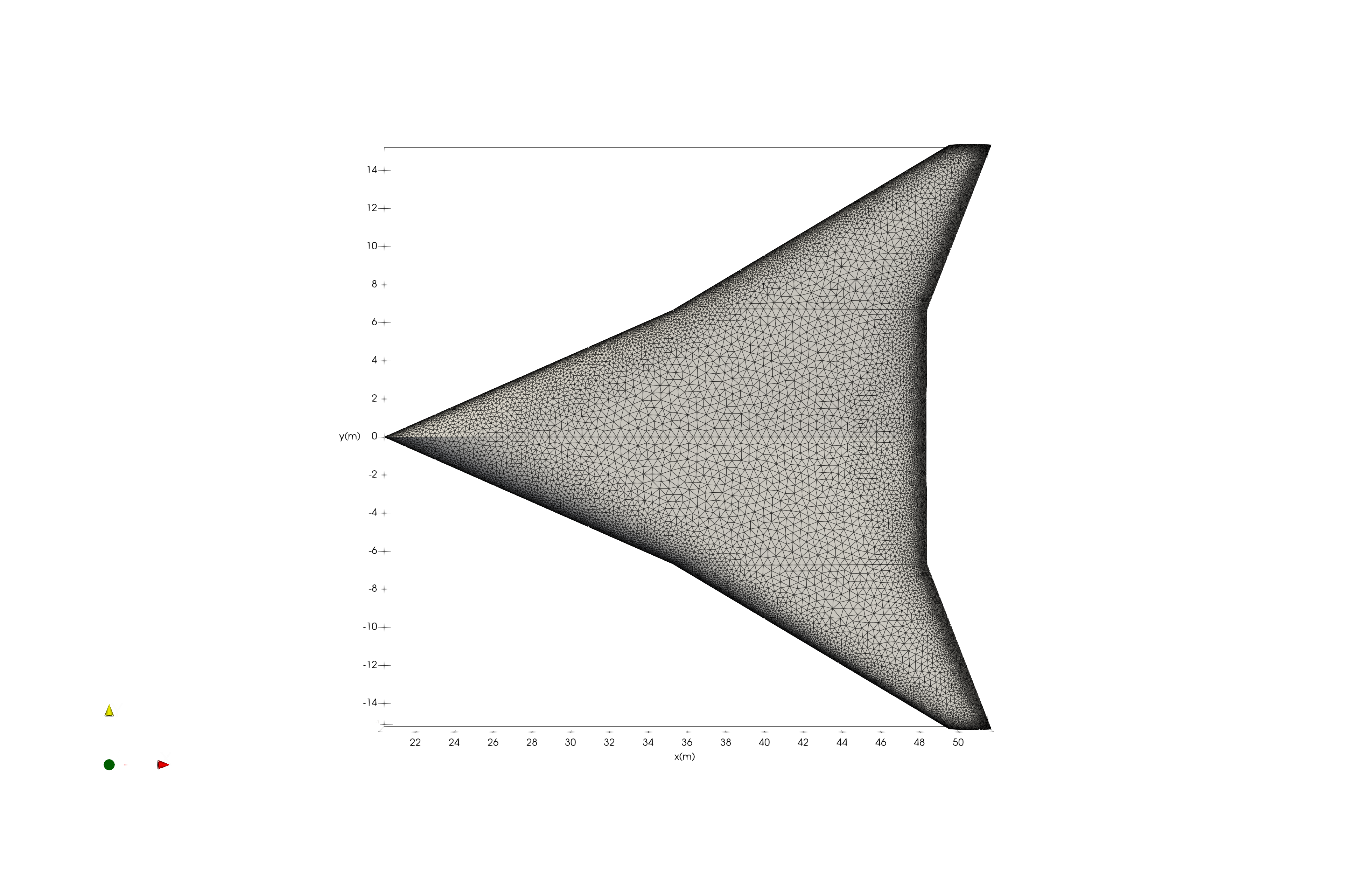}
    \end{minipage}
    \hfill
    \begin{minipage}{0.33\textwidth}
        \centering
        \textbf{(e) Level 5} \\[1ex]
        \includegraphics[width=\textwidth, trim={20cm 0cm 20cm 0cm}, clip]{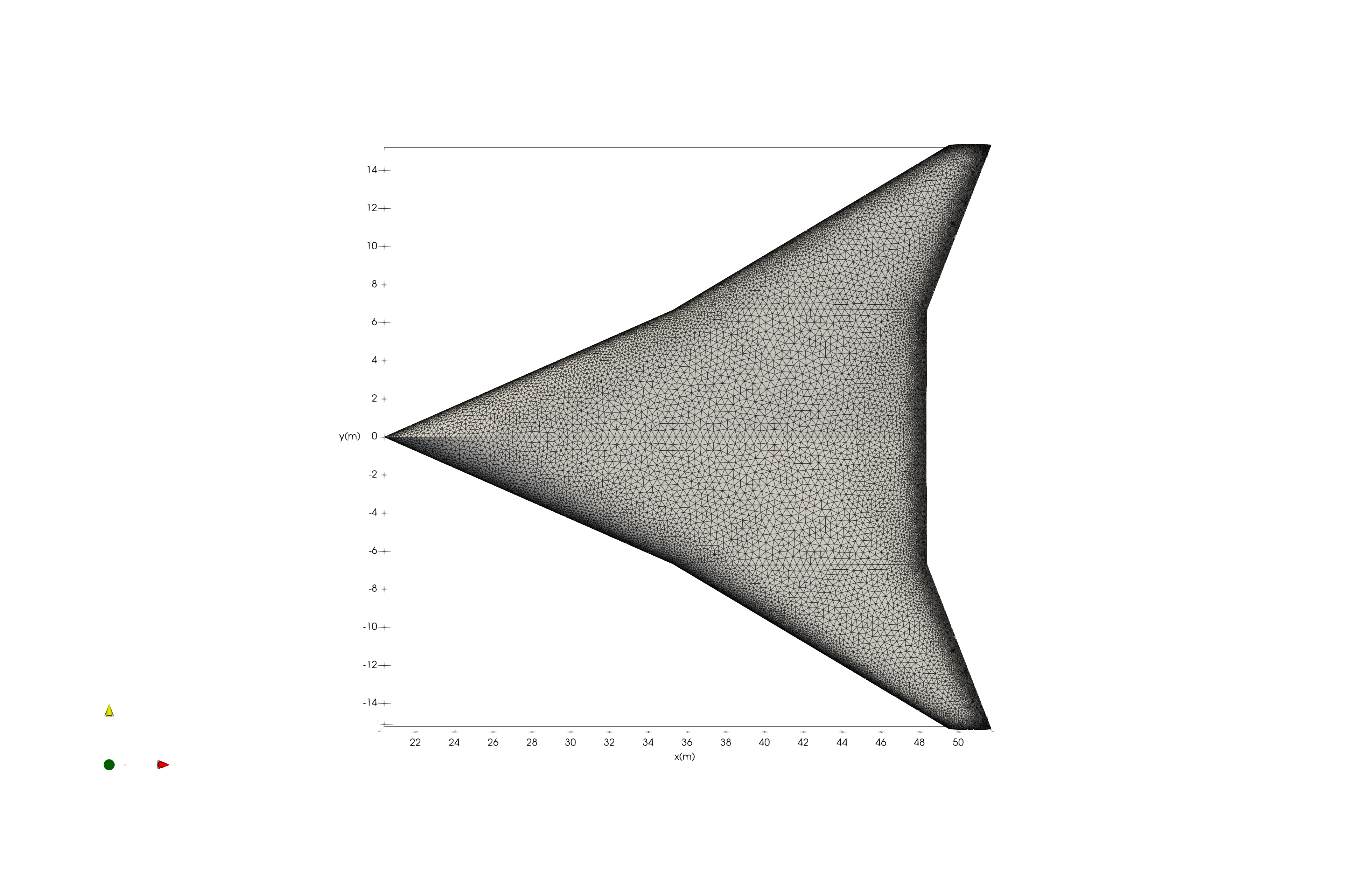}
    \end{minipage}
    \hfill
    \begin{minipage}{0.33\textwidth}
        \centering
        \textbf{(f) Level 6} \\[1ex]
        \includegraphics[width=\textwidth, trim={20cm 0cm 20cm 0cm}, clip]{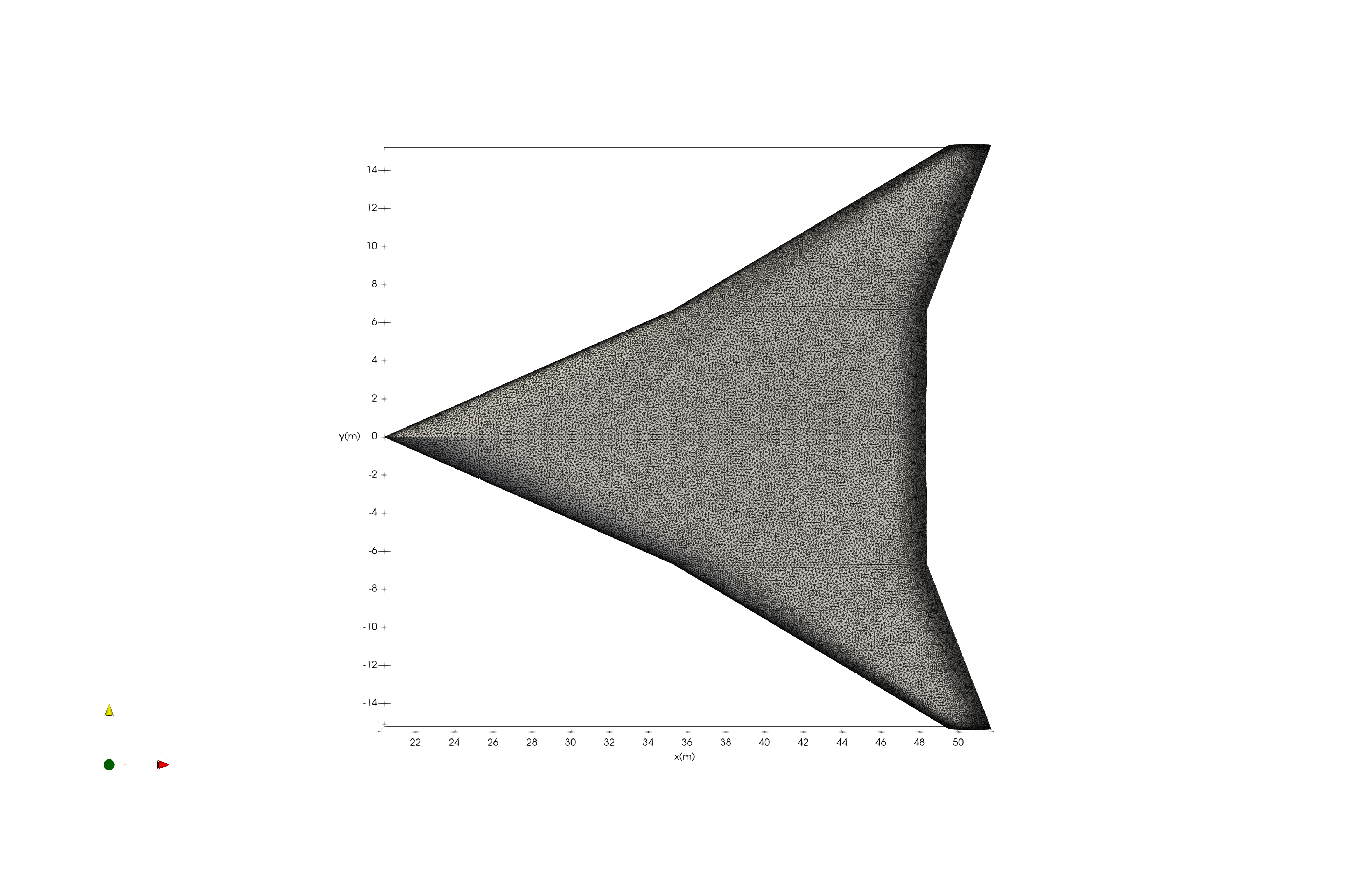}
    \end{minipage}

    \caption{Surface meshes of a representative geometry used for the mesh sizing study.}
    \label{fig:mesh-sizing}
\end{figure}

Several parameters control the mesh properties of \texttt{OpenVSP}'s \texttt{ComputeCFDMesh} surface meshes. These parameters include source definitions, the maximum mesh-edge length (enumerator \texttt{CFD\_MAX\_EDGE\_LEN}), the maximum edge-growth ratio (\texttt{CFD\_GROWTH\_RATIO}), and 3D rigorous smoothing and size-field limiting using nearest-neighbor matching of adjacent mesh elements (\texttt{CFD\_LIMIT\_GROWTH\_FLAG}). For the source definition, we add a trailing-edge source (\texttt{WLineSource}) with an influence radius of 0.1 \ac{MAC} and a mesh-edge-length target of 0.003 \ac{MAC}, in addition to the default sources at the leading edge and wing tips. The added trailing-edge refinement improved the volume-mesh quality near the trailing edge, and the refinement size is determined based on engineering judgment. The maximum edge-growth ratio is set to $1.2$, and 3D rigorous smoothing is enabled. We control the surface-mesh density by adjusting the maximum mesh-edge length. Six levels of mesh refinement are defined, with maximum edge lengths of $0.1, 0.05, 0.0375, 0.025, 0.02$, and $0.01$ \ac{MAC}.  We select a representative geometry from the dataset to determine a mesh configuration that is applied consistently across the full dataset. These surface meshes are shown in Figure~\ref{fig:mesh-sizing}. As expected, increasing the mesh-refinement level increases the mesh density. The corresponding surface cell element counts are $1.01\times10^5$, $1.05\times10^5$, $1.08\times10^5$, $1.18\times10^5$, $1.25\times10^5$, and $1.82\times10^5$ from Level~1 to Level~6 respectively.

The volume mesh is created from the exported surface mesh using a modified implementation of the \texttt{VSP2CFD} code \cite{vsp2cfd}. The primary parameters that control the volume-mesh properties are the maximum number of anisotropic \texttt{T-Rex} boundary-layer layers  (\texttt{TRexMaximumLayers}), the first layer thickness (\texttt{TRexCondition-setSpacing}), and the boundary decay rate of the unstructured mesh around wall surfaces (\texttt{BoundaryDecay}). For all mesh levels, the boundary decay rate is set to $0.85$. The first layer thickness values for the corresponding surface meshes from Level~1 to Level~6 are $6\times10^{-5}$, $2\times10^{-5}$, $9\times10^{-6}$, $7\times10^{-6}$, $6\times10^{-6}$, and $6\times10^{-6}$ (m), respectively. The maximum number of anisotropic layers is set to $25$, $35$, $40$, $50$, $45$, and $50$ layers from Level 1 to Level 6. As a result, the total volume cell counts are $5.32$, $6.13$, $6.57$, $6.96$, $7.54$, and $10.72$ million cells, respectively. The volume-mesh sizing study focuses primarily on identifying a proper boundary-layer mesh setup that allows the SA-R turbulence model \cite{Dacles-mariani-SARC} to capture the boundary-layer flow correctly, thereby yielding accurate aerodynamic and flow predictions. Table~\ref{tab:mesh-stats} summarizes the mesh configurations used in this study.

\begin{table}[h]
    \centering
    \begin{tabular}{lccc} \hline \hline
        \textbf{Level} & \textbf{Surface Elements} & \textbf{Volume Elements} & $\bar{y^+}$  \\\hline 
            1 & $1.01\times10^5$ & $5.32\times10^6$ & 6.93\\
            2 & $1.05\times10^5$ & $6.13\times10^6$ & 3.50\\
            3 & $1.08\times10^5$ & $6.57\times10^6$ & 1.98\\
            4 & $1.18\times10^5$ & $6.96\times10^6$ & 1.66\\
            5 & $1.25\times10^5$ & $7.54\times10^6$ & 1.35\\
            6 & $1.82\times10^5$ & $1.07\times10^7$ & 1.32\\
         \hline \hline
    \end{tabular}
    \caption{Mesh statistics across refinement levels.}
    \label{tab:mesh-stats}
\end{table}

\begin{figure}[h!]
    \begin{minipage}{0.49\textwidth}
        \centering
        \textbf{(a) $C_D$}
        \includegraphics[width=\textwidth]{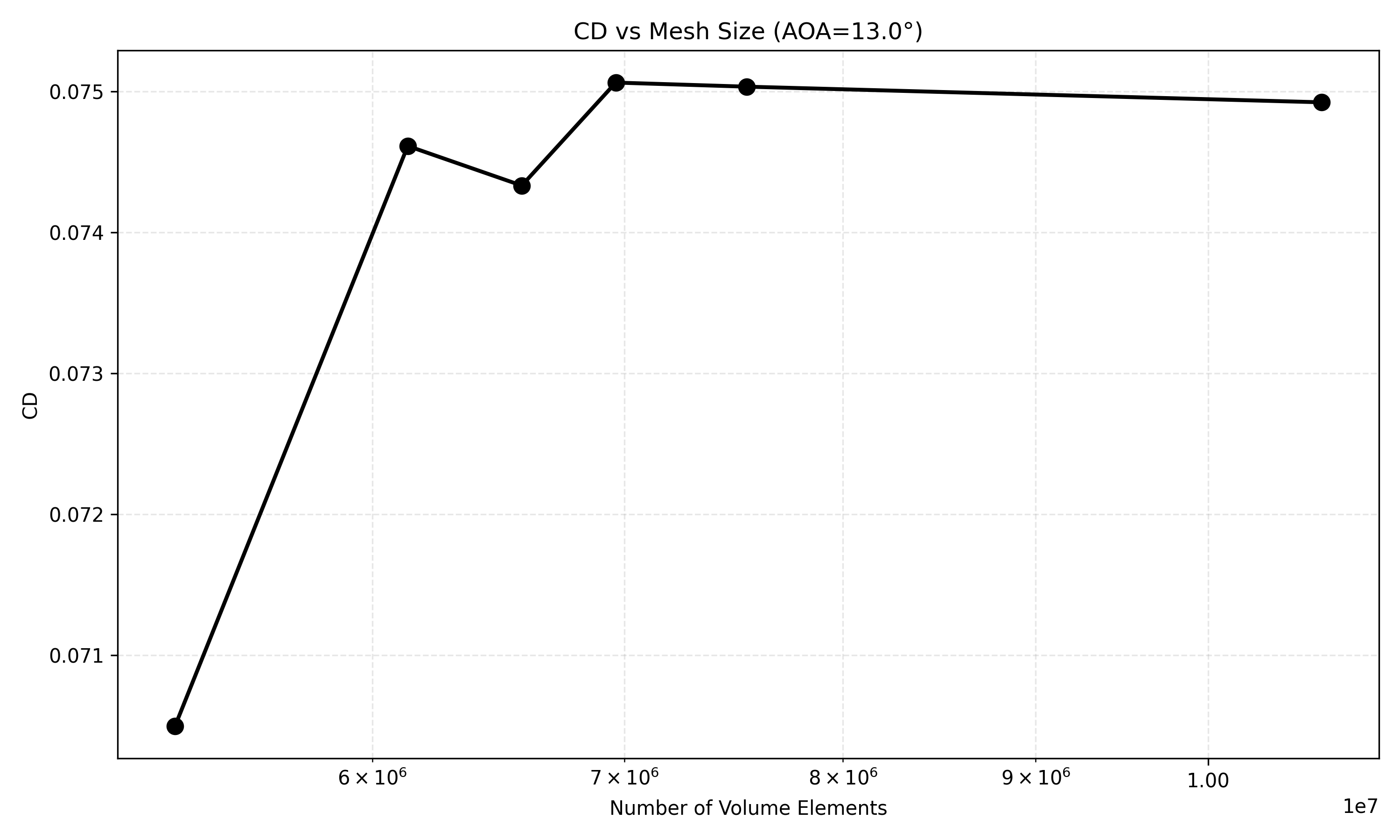}
    \end{minipage}
    \hfill
    \begin{minipage}{0.49\textwidth}
        \centering
        \textbf{(b) $C_M$}
        \includegraphics[width=\textwidth]{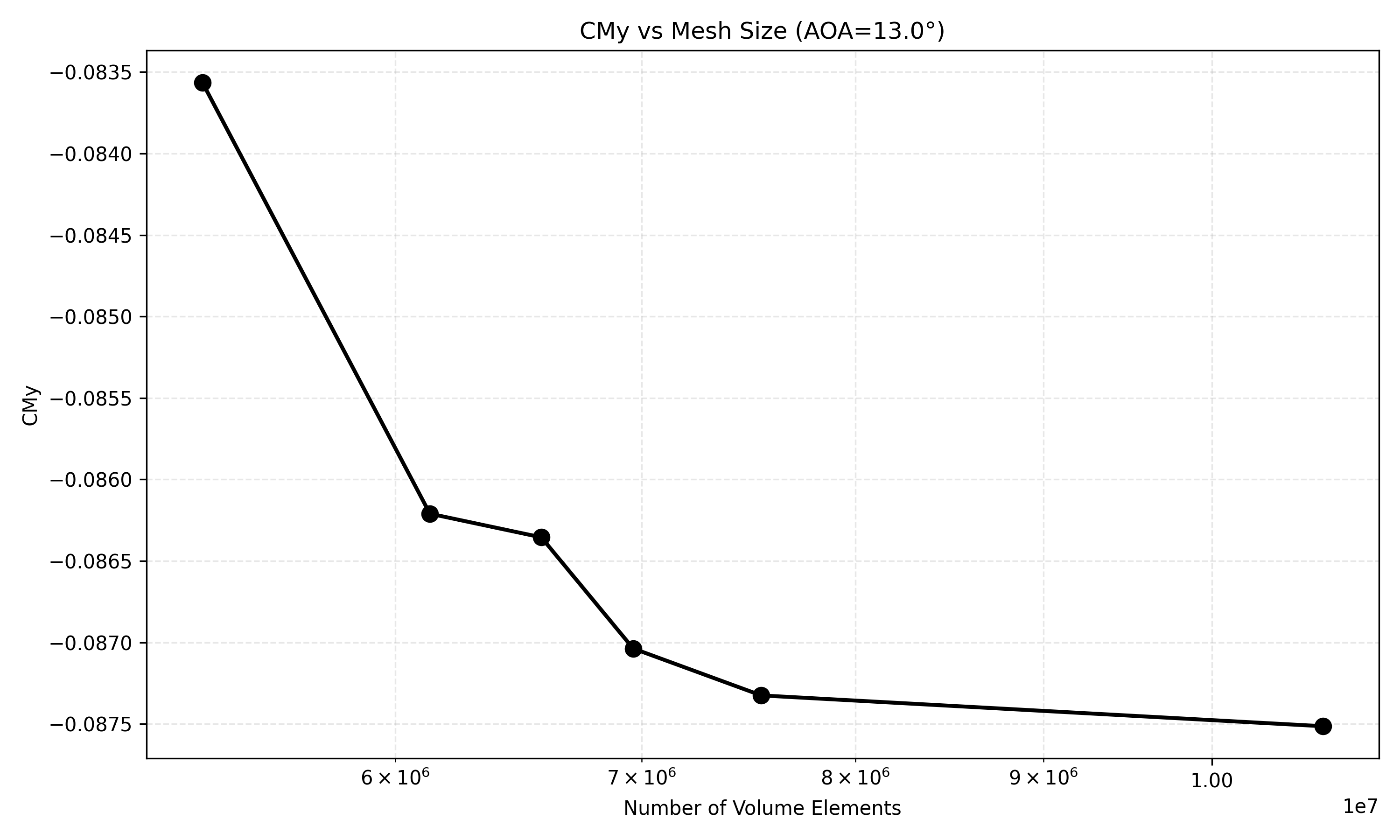}
    \end{minipage}
    \caption{Mesh sizing study: convergence of $C_D$ and $C_M$ with respect to mesh size.}
    \label{fig:mesh-sizing-convergence}
\end{figure}

A mesh convergence study is conducted by running \ac{CFD} simulations with identical freestream conditions and solver configurations across all mesh levels. We use the representative geometry at Mach 0.3 \ac{AOA} $13^\circ$ flow and monitor the convergence of $C_D$ and $C_M$, as well as the resultant average $\bar{y^+}$. We define the area-averaged $y^+$ as:
\begin{equation}
    \bar{y^+} = \frac{\int_S{y^+dA}}{ \int_SdA},
\end{equation}
and the resulting values for each mesh level are reported in Table~\ref{tab:mesh-stats}. For the SA-RC turbulence model without wall functions, we target $\bar{y^+} \approx 1$, which is achieved by selecting mesh level 4 and beyond. The convergence behavior for $C_D$ and $C_M$ is shown in Figure~\ref{fig:mesh-sizing-convergence}. It is observed that the Level~4 mesh with $6.96\times10^6$ volume cells achieves a difference within $1\%$ of the finest grid used in this study while maintaining a manageable computation cost for large dataset generation. Therefore, this mesh configuration is selected and used for all subsequent configurations' meshing.

\subsubsection{CFD Solver Settings}

The fluid domain consists of a freestream far-field with a diameter of 40 \ac{MAC} being 630 (m), centered at the wing apex, and the wing surface. The far-field boundary is assigned with a far-field boundary condition, while the wing surface is assigned with adiabatic wall (no-slip) condition. Freestream condition is specified as standard air, with gas constant of $287.06$, at $285$ (K), and static pressure of $7.18\times 10^4$ (Pa). The air has a constant heat capacity ratio of $1.4$. Sutherland viscosity model is used, with a reference viscosity of $1.715\times 10^{-5} (\text{Ns}/\text{m}^2)$ \cite{Sutherland01121893}, and a constant laminar Prandtl number of $0.72$ and turbulent Prandtl number of 0.9. The reference Reynolds number is based on average the \ac{MAC} of $16$ (m), yielding $Re_{ref} = 8.04\times10^7$. The aerodynamic coefficients are computed using a specific moment origin on the body and reference area defined according to the sizing rules in \texttt{SUAVE}. These configuration specific reference values are included in the dataset repository.

Spatial discretization employs the Jameson-Schmidt-Turkel (JST) scheme \cite{jst} for the flow equations with Green-Gauss gradient reconstruction, using a JST sensor tuned with coefficients (0.5, 0.02). The turbulence equation is discretized using a scalar upwind scheme without MUSCL reconstruction. Temporal advancement uses implicit Euler for both flow and turbulence, with an adaptive Courant-Friedrichs-Lewy (CFL) number. The linear systems are solved by FGMRES with ILU preconditioning and a residual tolerance of $10^{-7}$. The convergence criterion is set to be $10^{-6}$ for the continuity equation, with an additional convergence criterion for for the drag coefficient $C_D$ requiring changes below $10^{-5}$ over $100$ Cauchy iterations. These convergence criteria are applied consistently throughout all runs in the dataset.

\subsubsection{CFD Results and Discussion}
\begin{figure}[h!]
    \centering

    \begin{minipage}{0.33\textwidth}
        \centering
        \textbf{(a) 050101-7LK} \\[1ex]
        \includegraphics[width=\textwidth]{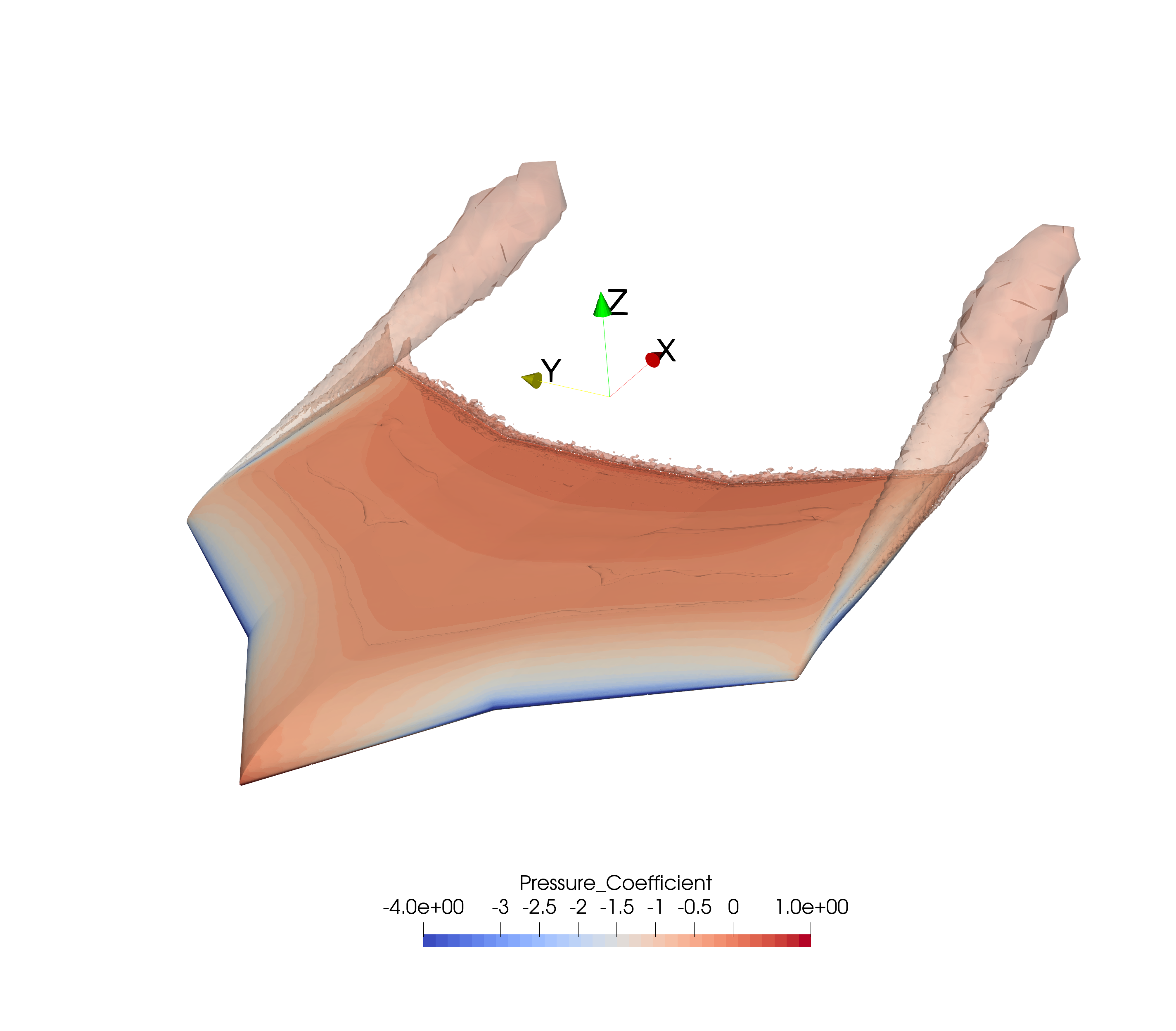}
    \end{minipage}
    \hfill
    \begin{minipage}{0.33\textwidth}
        \centering
        \textbf{(b) 09190107DL} \\[1ex]
        \includegraphics[width=\textwidth]{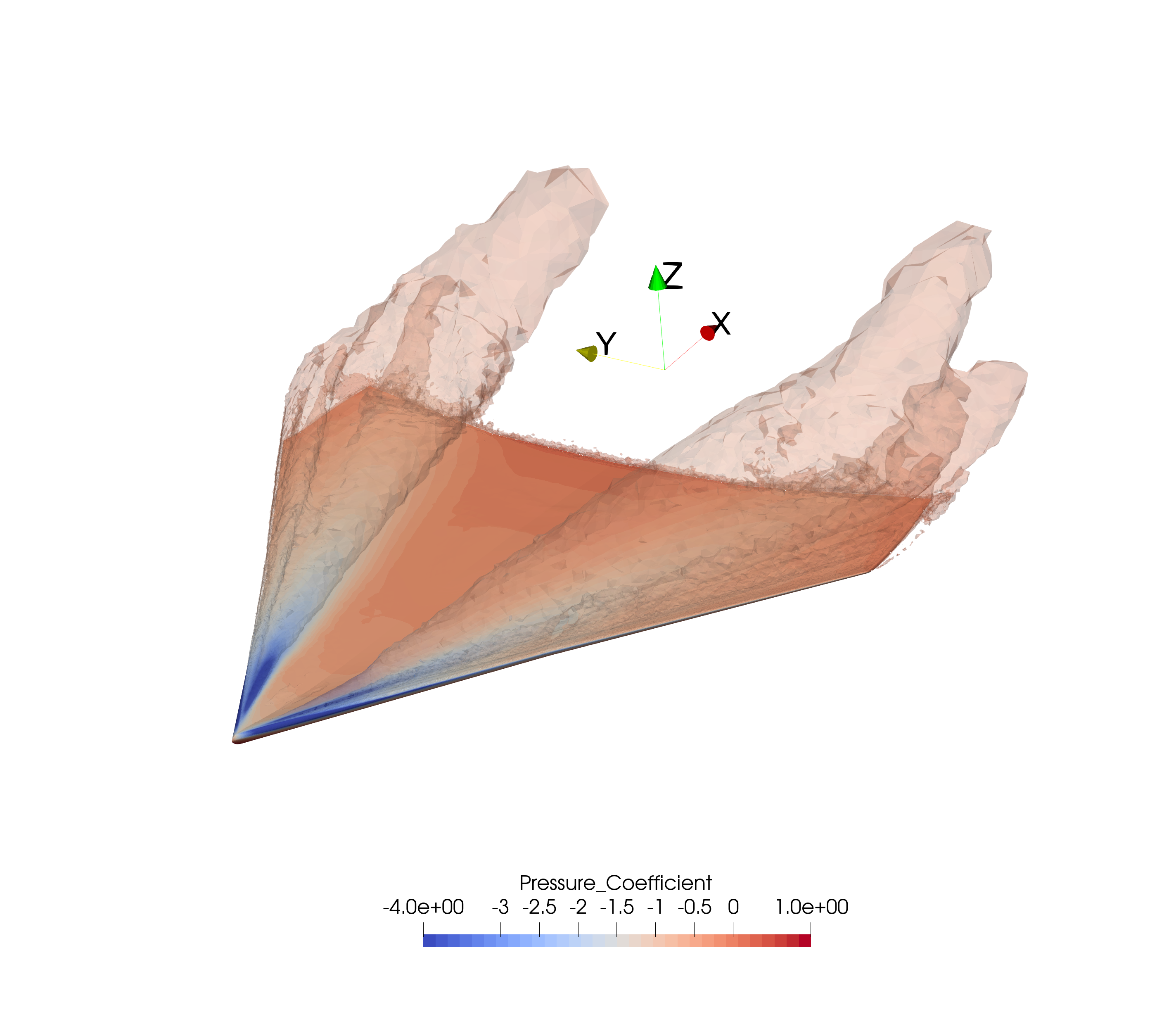}
    \end{minipage}
    \hfill
    \begin{minipage}{0.33\textwidth}
        \centering
        \textbf{(c) 171801-1NW} \\[1ex]
        \includegraphics[width=\textwidth]{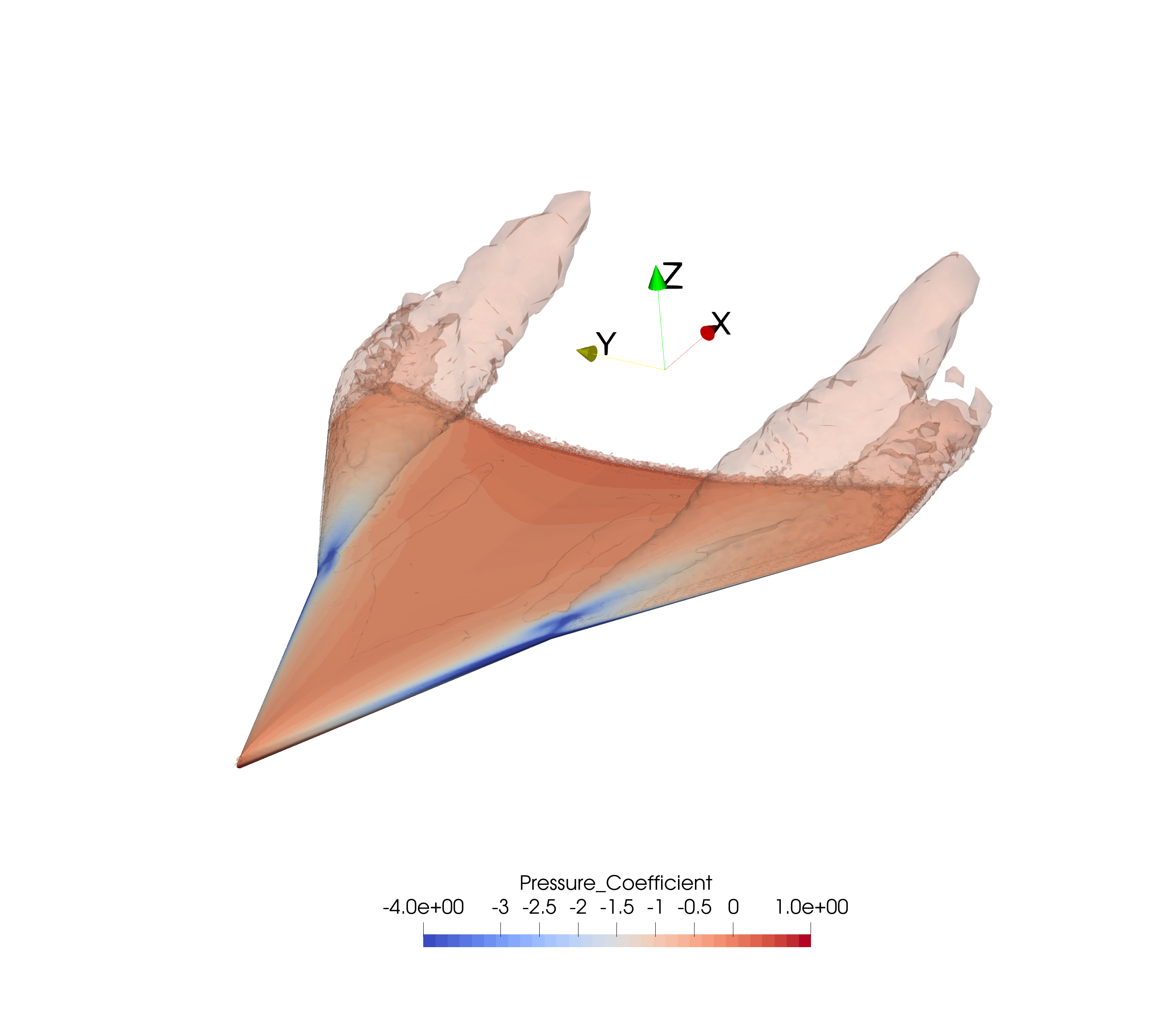}
    \end{minipage}
    \caption{Pressure coefficient ($C_P$) visualizations and Q-criterion iso-surfaces at $Q=100$ for different vehicle configurations, illustrating variations in geometry and flow states. }
    \label{fig:dataset-configurations}
\end{figure}

\ac{CFD} runs are conducted using the Stanford Research Computing Center (SRCC) facility with a 32-core 2.4 GHz partition. The total computational cost of the dataset is $310,835$ CPU-hours for $2,448$ snapshots. For illustration of the variety of flow conditions represented in the dataset, Figure~\ref{fig:dataset-configurations} shows three representative configurations (call signs as subtitles) at \ac{AOA} of $15^\circ$. These geometries are selected from the holdout test set, spanning the platform types ranging from trapezoid wings, cropped Delta wings, to double-delta wings. The wing surfaces are colored by $C_P$, and Q-criterion iso-surfaces of 100 are added to illustrate the vortex sheets. The variation of geometries leads to drastically different surface pressure distributions and primary vortex structures, enriching the geometric and flow features for surrogate modeling of flow prediction of the double-delta wing family.

Among the $2,448$ \ac{CFD} simulations, $57$ runs failed to converge to the specified residual target using the numerical scheme described above. These runs correspond primarily to highly cambered geometries at higher \acp{AOA} above $16^\circ$. The convergence failure arises from the unsteady nature of critically separated flow, for which the selected physics models and numerical scheme cannot recover a steady solution at the residual target. We believe it is more valuable to provide solutions with consistently converged results, even at a lower fidelity, rather than removing these snapshots. As a result, we lowered the order of the convective scheme and adopted AUSM \cite{liou2003ausm} for these runs. The configuration files contained in the dataset annotate which convective scheme is used for each run. Future extensions of the dataset may consider using higher-fidelity methods, such as detached eddy simulation, for these runs.

\subsection{Data Availability}
\label{sec:dataset-availability}

The dataset is hosted on our Huggingface repository at https://huggingface.co/datasets/yirens/double-delta-aero under GNU Lesser General Public License v2.1. The repository includes all \ac{CFD} results, including the \texttt{SU2} configuration files, surface and volume solutions in \texttt{vtk} format, convergence history files, and integrated aerodynamic loading summary files. The \ac{VLM} solutions are provided in \texttt{json} format. All files are organized and indexed by vehicle call sign for ease of access.

\section{Data Size Scaling Relationship}
\label{sec:scalingtest}

Neural scaling laws quantitatively describe how model performance improves with increasing training scale. The training scale comprises three major components: data scaling (how performance changes with increasing training dataset size), model scaling (how performance changes with increasing model size), and computation scaling (how model performance changes with increasing computational budget). Foundational work was carried out by \citeauthor{hestness2017deeplearningscalingpredictable} \citeauthor{rosenfeld2019constructivepredictiongeneralizationerror}, \citeauthor{kaplan2020scalinglawsneurallanguage}, and \citeauthor{hoffmann2022trainingcomputeoptimallargelanguage}, focusing on training large language models  \cite{hestness2017deeplearningscalingpredictable, rosenfeld2019constructivepredictiongeneralizationerror, kaplan2020scalinglawsneurallanguage, hoffmann2022trainingcomputeoptimallargelanguage}. Together, the studies show that model performance under a given computational budget $C$, when measured by an appropriate loss metric, follows a power-law relationship with training dataset size and model size:

\begin{equation}
    \epsilon = a_1 N^{-\alpha} + a_2 D^{-\beta} + \epsilon_0,
    \label{eqn:scaling}
\end{equation}
where $\epsilon$ is the test loss, $N$ is the model size, $D$ is the dataset size, and $\epsilon_0$ is the irreducible error. Coefficients $a_1$, $a_2$, $\alpha$, $\beta$, and $\epsilon_0$ are fitting parameters. Such formulation is applicable to GNNs, as indicated by previous study  \cite{liu2024neuralscalinglawsgraphs}.

We define $\epsilon$ as the \ac{MSE} evaluated on the holdout test set. Recall that \texttt{MF-VortexNet} learns a mapping $\mathcal{F}$ that takes a low-fidelity pressure--stream-condition--geometry state, represented as a graph $G$, and produces a predicted pressure distribution over the lattice, $C_p^{\text{pred}}$, that approximates the ground-truth pressure distribution $C_p^{\text{gt}}$ obtained from high-fidelity simulations. The test loss is then defined as

\begin{equation}
    \epsilon = \text{MSE}(C_p^{\text{gt}}, C_p^{\text{pred}}) .
\end{equation}

The remainder of this section outlines the procedure used to identify the scaling-law coefficients based on the test error $\epsilon$, with the results interpreted as preliminary.

\subsection{Experiment Setups}
\label{ssec:scaling-exp-setups}

The empirical scaling experiment aims to estimate the exponent coefficients in Equation~\ref{eqn:scaling}, enabling us to quantify how prediction accuracy scales with training data size and to determine the data requirements necessary to achieve a target accuracy. The experiment comprises three components: the training data, the model, and the training configurations.

\subsubsection{Data}
\label{sssec:scaling-exp-setups-data}

We define the training data scaling under investigation as the increase in the number of snapshots resulting from higher sampling density over the a defined geometric design space. As summarized in Table~\ref{tab:dataset-levels}, using Saltelli sampling, the geometric configurations in the training set (described in Section~\ref{sec:dataset}) are subdivided into six levels. From Level~1 to Level~6, the training set includes a total of 8, 16, 32, 64, 128, and 256 geometric configurations, respectively. For each level, flow snapshots at five \acp{AOA} are selected at $11^\circ$, $13^\circ$, $15^\circ$, $17^\circ$, and $19^\circ$. Together, our training data scaling involves $D$ values of $40, 80, 160, 320, 640$, and $1280$ graphs, respectively. Each graph provides a multi-fidelity representation of a geometric configuration and its associated flow state.

The decision to include only five \acp{AOA} per geometry, rather than all available \acp{AOA} in the dataset, is motivated by the need to balance dataset size and model capacity under tight computational budget. Because the primary objective of the data scaling study is to determine the geometric sampling density required to train \texttt{MF-VortexNet}, we prioritize scaling the number of geometries rather than the number of flow snapshots per geometry. As discussed by \citeauthor{hoffmann2022trainingcomputeoptimallargelanguage}, dataset size, model size, and computational budget should be scaled proportionally \cite{hoffmann2022trainingcomputeoptimallargelanguage}. Subsampling the available \acp{AOA} therefore enables training with smaller models at lower computational cost, while preserving the qualitative empirical scaling trends.

An additional consideration concerns how to define the dataset size $D$ for a collection of graphs. Prior studies across different domains have adopted different definitions, including the volume of digital information measured in terabytes \cite{li2025scalinglawsgraphneural}, the number of individual graphs \cite{sypetkowski2024on}, and the total number of edges \cite{liu2024neuralscalinglawsgraphs}. Consequently, there is currently no standardized definition of $D$ in graph-based setting, and its interpretation remains domain-specific. In this work, we define $D$ as the total number of graphs in the dataset. This definition is aligned with the downstream design application, where designers seek to understand how densely the design space needs to be sampled and how many high-fidelity simulations are required to train a surrogate model with sufficient accuracy. Such a definition directly links to the number of simulation runs required and is thus more convenient for budget planning.

Besides the training set, we constructed the test set using the sixteen holdout test geometries, as described in Section~\ref{sec:dataset}. To maintain the consistency with the training set in flow snapshot selection, we include include snapshots at \acp{AOA} of $11^\circ$, $13^\circ$, $15^\circ$, $17^\circ$, and $19^\circ$. In total, the test set comprises $80$ graphs. The test set is relatively small. However, considering the even geometric distribution shown in Figure~\ref{fig:dspace-pairplot}, we believe the test set contains adequate variability representative of the training set and is suitable for use in test loss evaluation.

\subsubsection{Model}
\label{sssec:scaling-exp-setups-model}

Prior work has shown that performance scaling depends critically on the relationship between dataset size and model capacity. In the under-parameterized or over-parameterized regimes, corresponding to a disproportionately small or large ratios of $N/D$, the test error no longer exhibits a power-law decay with respect to $D$, a phenomenon referred to as model scaling collapse \cite{hoffmann2022trainingcomputeoptimallargelanguage, liu2024neuralscalinglawsgraphs}. To avoid inconclusive results derived from a single fixed-capacity model, we repeat the scaling experiment using four models with model sizes ranging from 0.10 million to 2.41 million trainable parameters.

The \texttt{MF-VortexNet} architecture has three hyper-parameters that are most relevant to model size \cite{Shen2025VortexNet}, including the width of the nodal encoding for hidden layers (\texttt{hidden\_feature}), the number of heads for multi-head attention (\texttt{heads}), and the total depth of the message passing blocks (\texttt{hops}). The \texttt{hops} parameter is related to the extent of nodal information propagation, and for the given $30\times32$ lattice, we determined that a \texttt{hops} of 10 is adequate for the information to propagate across all nodes in a graph. This hyper-parameter is therefore held constant across models. To change the model size, we alter \texttt{heads} and \texttt{hidden\_feature}. Specifically, increasing \texttt{heads} and \texttt{hidden\_feature} increases the model size while widening the model. The resulting models created, along with their numbers of trainable parameter size ($N$), are presented in Table~\ref{tab:model-sizes}.

\begin{table}[h!]
    \centering
    \begin{tabular}{lccc} \hline \hline
        \textbf{Level} & \textbf{\texttt{hidden\_feature}} & \textbf{\texttt{heads}}  & \textbf{Model Size ($N$)}\\\hline 
            Mini & 13 & 5 & $0.103\times10^6$   \\
            Small & 27 & 5 &  $0.538\times10^6$ \\
            Medium & 37 & 5 & $1.144\times10^6$  \\
            Large & 45 & 6 &  $2.411\times10^6$  \\
         \hline \hline
    \end{tabular}
    \caption{Model sizes at four levels.}
    \label{tab:model-sizes}
\end{table}

Once the model size is defined, additional hyper-parameters related to regularization and training must be specified \cite{Shen2025VortexNet}. Assigning a single fixed set of hyper-parameters across all models is not feasible, as models of different sizes require different training dynamics and regularization strategies to achieve convergence and target performance accuracy. Instead, we promote cross-model comparability by adopting a consistent hyper-parameter selection procedure that yields ``equally optimal'' configurations for each model.

For each model size, we employ \texttt{Optuna} \cite{optuna_2019} to optimize key training and regularization hyper-parameters, including the learning rate and weight decay of the ADAM optimizer, as well as the dropout ratio and skip-connection bandwidth (\texttt{Alpha}). Hyper-parameter optimization is performed using a subset of the Level~6 training set, constructed by randomly sampling $2048$ snapshots, thereby ensuring that no information from the holdout test set is leaked during model selection. This dataset is further randomly split into an 80\%-20\% training-validation partition. For each model, the training configuration that achieves the lowest validation \ac{MSE} is adopted as the training hyper-parameters.

To ensure fair comparison across model capacities, an identical computational budget, as discussed in the following section, is allocated to all models. We conduct a maximum of 35 training runs during hyper-parameter optimization for each model, with the first ten trials used as ``warm start'' evaluations for \texttt{Optuna}. All other \texttt{Optuna} optimization settings are left at their default values.

The \ac{MSE} values of the optimized models are reported in Table~\ref{tab:model-hpoptimized}. Despite differences in model sizes and training configurations, comparable test \ac{MSE} values are achieved. The remaining hyper-parameters are reported to ensure reproducibility. These hyper-parameters are held fixed for each model across all data scaling runs.

\begin{table}[h!]
    \centering
    \begin{tabular}{lccccc} \hline \hline
        \textbf{Level} & \textbf{Test \ac{MSE}} & \textbf{Learning Rate}  & \textbf{Decay} & \textbf{Dropout} & \textbf{\texttt{Alpha}} \\\hline 
            Mini & $5.97\times10^{-2}$ & $1.14\times10^{-3}$ & $2.81\times10^{-6}$ & $5.57\times10^{-2}$ & $4.50\times10^{-1}$\\
            
            Small & $5.66\times10^{-2}$ & $6.59\times10^{-3}$ & $1.54\times10^{-6}$ & $1.22\times10^{-1}$ & $2.38\times10^{-1}$ \\
            
            Medium & $5.21\times10^{-2}$ & $1.60\times10^{-3}$ & $1.98\times10^{-6}$ & $5.19\times10^{-2}$ & $4.10\times10^{-1}$\\
            
            Large & $5.88\times10^{-2}$ & $6.02\times10^{-4}$ & $3.28\times10^{-6}$ & $1.82\times10^{-1}$ & $2.77\times10^{-1}$ \\
         \hline \hline
    \end{tabular}
    \caption{Additional training hyper-parameters after optimization for all models.}
    \label{tab:model-hpoptimized}
\end{table}

\subsubsection{Training}
\label{sssec:scaling-exp-setups-trainingprotocol}

To fit the neural scaling law coefficients in Equation~\ref{eqn:scaling}, we conduct a controlled data-scaling experiment for each model listed in Table~\ref{tab:model-sizes}. The training-dataset size $D$ is systematically varied, as discussed in Section~\ref{sssec:scaling-exp-setups-data}, while keeping the training procedure fixed. This procedure is designed to isolate the effect of data scaling from variations in model capacity or computational cost.

When performing the $D$ scaling using datasets at six levels, as discussed in Section~\ref{sssec:scaling-exp-setups-data}, and across four model sizes $N$, careful design of the training scheme is needed such that, as a result of the experiment, the fitted $\beta$ represents the training efficiency gain resulting from increased geometric sampling density, rather than being a consequence of increased computation. As an experimental design choice, we fix the maximum computational budget $C^*$ available for all training runs across all $D$ and $N$. $C^*$ is defined as the maximum number of weights update steps available for each training fold, and is set to $2000$. Four folds are used to train the model; in each fold, a 75\%--25\% training-to-validation data partitioning is used. Adopting k-fold cross-validation reduces the risk of model overfitting and is a critical component of weight scheduling for physics-informed loss integration \cite{Shen2025VortexNet}. In addition to fixing $C^*$, an early stopping patience of $50$ epochs is used across training runs to prevent model overfitting. The early stopping mechanism terminates the training of the current fold when no improvement in validation loss is observed over the preceding $50$ epochs. After examining the validation loss histories across epochs for all runs, we did not observe typical overfitting behavior, such as validation loss increases toward the end of training, except for the large model trained on the Level~1 dataset, which, due to the limited amount of validation data, is more prone to overfitting, supporting the appropriateness of the chosen early stopping patience.  

Setting a fixed maximum number of weight-update iterations introduces an additional challenge for dataset-size scaling. In standard mini-batch gradient descent, each batch corresponds to a single weight-update step; for a training dataset of size $D_t$ and batch size $B$, a total of $D_t/B$ weight updates are performed per epoch. Since $B$ is constrained by hardware memory and is fixed at $128$, the total number of weight updates scales with $D_t$, introducing an inconsistency when isolating the effect of dataset size on model performance. Moreover, prior work suggests that the optimal computational budget $C$ does not necessarily scale linearly with $D$ \cite{hoffmann2022trainingcomputeoptimallargelanguage}, and the scope of the present study is insufficient to identify the optimal scaling relationship between $C$ and $D$. Consequently, during the experiment we found the model hyper-parameters, as discussed in Section~\ref{sssec:scaling-exp-setups-model}, are sensitive to $C$, such that designing a $C$ allocation scheme across model sizes and data sizes for calibrating the performance becomes difficult. An imprudent $C$ scheduling may result in biased power-law observations.

To address this issue and isolate error scaling with respect to $D$ alone, we fix the number of weight-update steps across dataset sizes by adopting a stochastic subsampling scheme. Specifically, at each epoch, we randomly draw $B$ samples from the training set and perform a full-batch gradient update. Because of the stochastic nature of this procedure, and given a sufficiently large number of epochs, the model is exposed to the entire training dataset in expectation, while the total computational budget $C$ remains constant across different dataset sizes. As a result, for each fixed model size, the observed scaling behavior can be attributed solely to changes in sample distribution introduced by $D$ scaling. Since the same number of flow conditions are selected at each level of the geometry set, this scaling behavior directly reflects gains in prediction accuracy arising from increased sampling density in the design space. The resultant $\epsilon$ versus $D$ scaling results are conceptually similar to the iso-FLOP experiments reported by \citeauthor{hoffmann2022trainingcomputeoptimallargelanguage} \cite{hoffmann2022trainingcomputeoptimallargelanguage}.

The remaining training setup follows \citeauthor{Shen2025VortexNet} \cite{Shen2025VortexNet}. For each model training at a particular model size $N$ and dataset size $D$, we repeat the training six times under randomized initial seeds for initial model weights and dataset partitioning. Trial-level \acp{MSE} are aggregated to report per-level mean, standard deviation, minimum, and maximum values to improve the robustness of the scaling exponent estimation. Least-squares fitting using \texttt{SciPy}'s \texttt{curve\_fit} function is used to fit the aggregated mean \acp{MSE} to the scaling law coefficients in Equation~\ref{eqn:scaling}. All model trainings are executed on a 40GB NVIDIA A100 GPU.

\subsection{Results and Discussion}
\label{ssec:scaling-results}

\begin{figure}[hbtp!]
    \centering
    \includegraphics[width=0.7\linewidth]{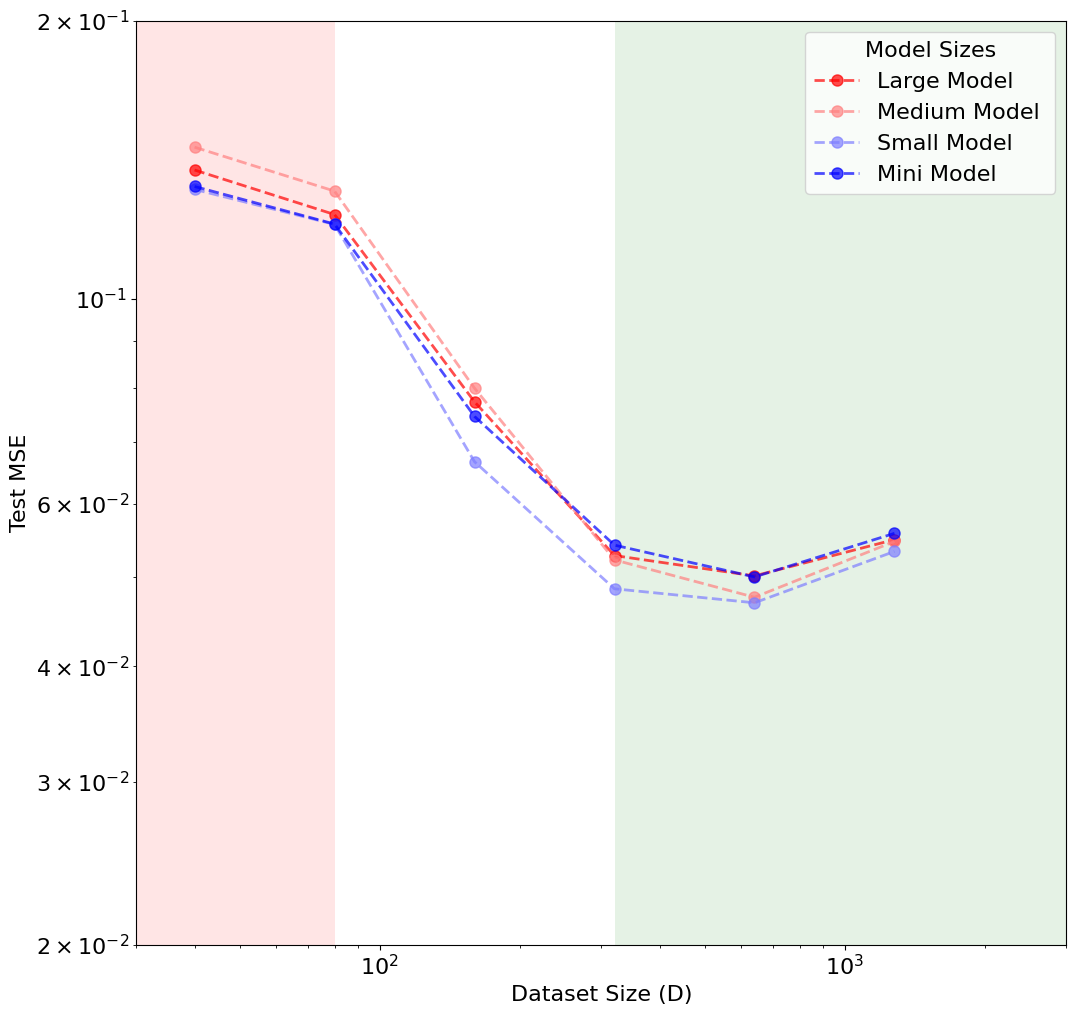}
    \caption{Scaling behavior of test \ac{MSE} as a function of the training set size $D$ for models of different sizes, where $D$ denotes the number of distinct geometry-flow snapshots represented as graphs in the dataset. }
    \label{fig:scaling-all-models}
\end{figure}

The mean test \ac{MSE}, evaluated on the holdout test set and averaged across all trails at each $D$, is plotted in Figure~\ref{fig:scaling-all-models}. Different data series represent scaling results from different models, from large to small. Overall, the test \ac{MSE} decreases as $D$ increases, although the rate of convergence varies across models. From the learning curves, three distinct learning regions can be identified. We highlight these regions by light red, white, and light green bands along the $D$ axis. Such zoning is consistent with the learning curve behavior described by \citeauthor{hestness2017deeplearningscalingpredictable} and \citeauthor{rosenfeld2019constructivepredictiongeneralizationerror} \cite{hestness2017deeplearningscalingpredictable, rosenfeld2019constructivepredictiongeneralizationerror}.

When the dataset is small (light red region), the learning curves exhibit behavior characteristic of the ``small data region''\cite{hestness2017deeplearningscalingpredictable}, in which model performance is limited by insufficient training data provided. Consequently, the model performs a best-guess prediction. For models investigated in this study, this region corresponds to $D\leq 80$. As $D$ increases beyond the ``small data region'', a ''power-law region'' \cite{hestness2017deeplearningscalingpredictable} emerges, such that increasing $D$ in this region helps the model to improve its prediction quality for generalization tasks, such as predictions in the holdout test set. Under the current model and training configuration, this region spans $80 \leq D \leq 320$, corresponding to geometry sets of $16$ to $64$ configurations.

For larger data sizes, at $D\geq 320$, an ``asymptotic error region'' is observed. We intentionally distinguish the terminology of this region from the ``irreducible error region'' described by \citeauthor{hestness2017deeplearningscalingpredictable} \cite{hestness2017deeplearningscalingpredictable}, as the error floor observed in Figure~\ref{fig:scaling-all-models} should not be interpreted as an inherent limitation of the dataset or as insufficient model expressivity. Instead, it arises from training under a fixed computational budget $C^*$, as discussed in Section~\ref{sssec:scaling-exp-setups-trainingprotocol}. To achieve fixed $C^*$ across different values of $D$s, subsampling is used. Under the training setup of a batch size of $128$, training data for $D\geq 170$ cannot fit into the full-batch gradient descent, and thus the training data, when viewed as an entire set, are underutilized in each weight update step. Although all graphs in the training set are likely to be used during training, each graph in larger datasets is less exposed to the optimizer, and hence the resulting models are less optimally trained. This under-training effect explains why the test \ac{MSE} for models trained at $D=1280$ tends to be higher than those trained at $D=640$. These trends are consistent with the iso-FLOP curves reported in previous literature \cite{hoffmann2022trainingcomputeoptimallargelanguage}, and this region should not be used for power-law fitting as the model training is bottlenecked by $C^*$.

\begin{figure}[hbtp!]
    \centering

    \begin{minipage}{0.41\textwidth}
        \centering
        \textbf{(a) Mini Model Scaling}
        \includegraphics[width=\textwidth]{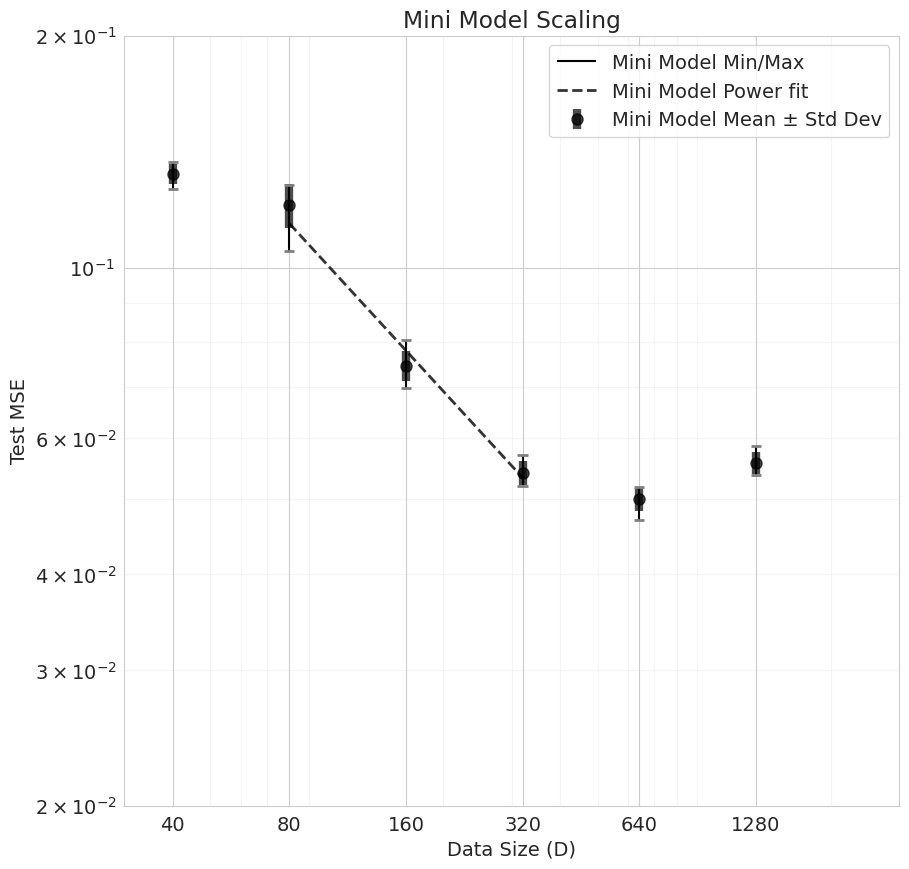}
    \end{minipage}
    \hfill
    \begin{minipage}{0.41\textwidth}
        \centering
        \textbf{(b) Small Model Scaling}
        \includegraphics[width=\textwidth]{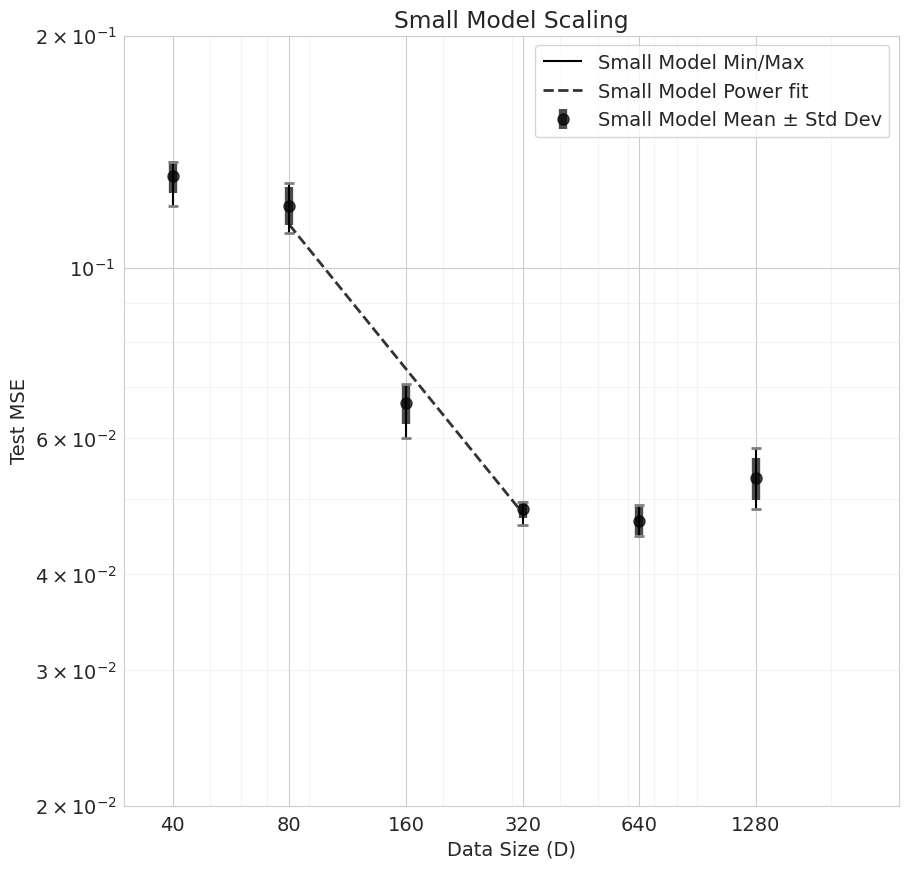}
    \end{minipage}
    
    \centering

    \begin{minipage}{0.41\textwidth}
        \centering
        \textbf{(c) Medium Model Scaling}
        \includegraphics[width=\textwidth]{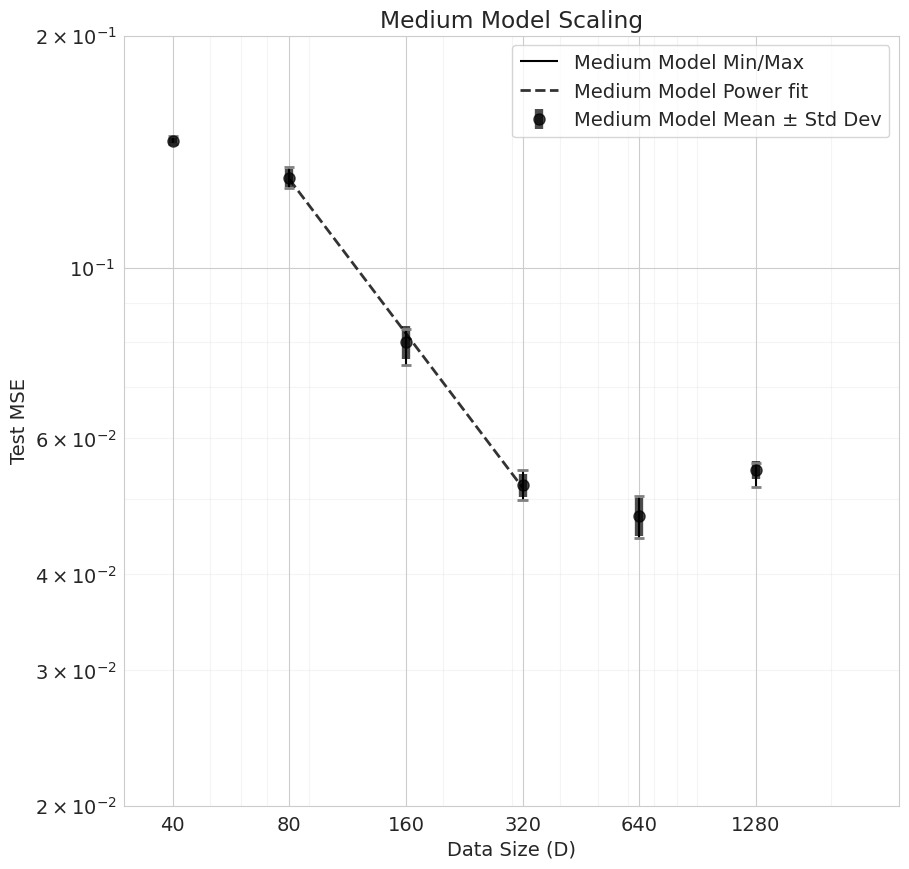}
    \end{minipage}
    \hfill
    \begin{minipage}{0.41\textwidth}
        \centering
        \textbf{(d) Large Model Scaling}
        \includegraphics[width=\textwidth]{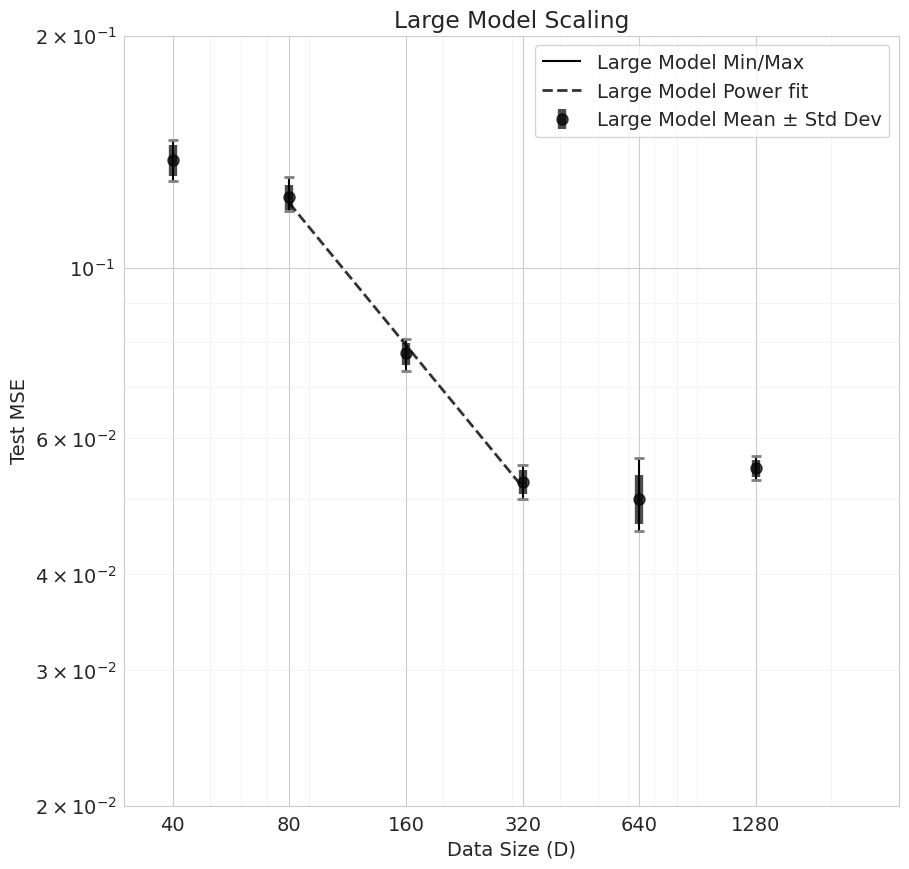}
    \end{minipage}

    \caption{Test \ac{MSE} scaling with $D$ for four models arranged from (a) to (d): Mini, Small, Medium, and Large, along with their log-log linear fit for the power-law region.}
    \label{fig:scaling-fitting}
\end{figure}

Among these regions, the power-law region provides information on data-size scaling. Figure~\ref{fig:scaling-fitting} presents the test \ac{MSE} as a function of $D$, together with the mean value, variability indicated by boxes spanning plus and minus one standard deviation, and the minimum and maximum values shown by whiskers in a box-plot style. The mean values are identical to those presented in Figure~\ref{fig:scaling-all-models}. It is observed that the larger models tends to have smaller coefficient of variation in the power-law region, indicating a more consistent performance across repeated trials. 

Within the power-law region, Equation~\ref{eqn:scaling} can be further reduced to,
\begin{equation}
    \epsilon' = a_2D^{-\beta},
    \label{eqn:power-logliner}
\end{equation} 
by absorbing the constant contributions from model size and irreducible error into the test loss, we focus on quantifying the scaling exponent $\beta$. To do so, we use \texttt{SciPy}'s \texttt{curve\_fit} function to perform a non-linear least squares fit of the data. The mean \ac{MSE} is treated as the dependent variable, and $D$ is treated as the independent variable, with the corresponding standard deviations used to represent the relative uncertainty of the \ac{MSE} measurements. The resulting power-law fits are shown as dashed lines in Figure~\ref{fig:scaling-fitting}(a)–(d).

From the power-law fitting, we estimate the scaling exponent $\beta$ to be $0.5484$, $0.6220$, $0.6654$, and $0.6128$ for the Mini, Small, Medium, and Large models, respectively. The corresponding coefficient $a_2$ are $1.2617$, $1.7354$, $2.4044$, and $1.7778$, respectively. The corresponding R-squared values are $0.9772$, $0.9652$, $0.9985$, and $0.9965$ respectively, indicating a high-quality fits for $\beta$ estimation.

These fitted power laws can then be used to answer the question of how much data is required to achieve a target prediction accuracy under the proposed training procedure and computational budget. Assuming that the minimum \acp{MSE} obtained from hyper-parameter optimization (Table~\ref{tab:model-hpoptimized}) are reasonable estimates of models' best performance under substantially larger dataset and are served as the loss target, we compute the optimal dataset size $D^*$ at the computational budget $C^*$ by inverting Equation~\ref{eqn:power-logliner}. The resulting values of $D^*$ are 261, 245, 317, and 261 for the Mini, Small, Medium, and Large models, respectively. These values correspond to geometry sets of 52, 49, 63, and 52 geometries, all of which are slightly smaller than the Level~4 dataset listed in Table~\ref{tab:dataset-levels}.

The size of the geometry set can be further mapped to the corresponding sampling density in the unit hypercube. To this end, we first fit a power law relating the average nearest-neighbor distance $h$, defined as the average Euclidean distance from each design sample to its nearest neighbor, to the number of geometry samples $G$ contained in a dataset, using the data in Table~\ref{tab:dataset-levels}. This fit yields $h = 0.62 G^{-0.15}$, from which the estimated $D^*$ values correspond to $h$ in the range of $0.333$ to $0.346$. The corresponding one-dimensional optimal sampling density can then be estimated for space-filling sampling algorithms by accounting for dimensional scaling. Based on this analysis, selecting approximately eight sampling points per dimension is sufficient to achieve optimal surrogate model performance under the allocated computational budget $C^*$.

The data requirement for training \texttt{MF-VortexNet} is considerably lower than those of other machine-learning surrogate models capable of field prediction \cite{Pfaff2021MeshGraphNets, ranade2025dominodecomposablemultiscaleiterative, alkin2025abuptscalingneuralcfd, Catalani2024, zhao-deeponet-2023}. Such models typically require hundreds of geometries to achieve geometric generalizability, along with thousands of flow simulation snapshots to achieve field prediction accuracy. We believe the efficiency in data scaling roots in \texttt{MF-VortexNet}'s design in operating effectively in small-data use cases. By incorporating physics-informed loss functions, scheduled training strategies, and relatively compact model architectures, prior studies \cite{Shen2025VortexNet, shen_2025_aviationGNNASO} have demonstrated that strong interpolation capability within the design space can be achieved using only 15 to 27 geometries with 189 to 600 flow snapshots. However, because the model remains relatively small compared to  large-capacity architectures \cite{ranade2025dominodecomposablemultiscaleiterative}, its representational capacity is inherently limited. As a result, this model size and training scheme constraint may prevent the surrogate from fully exploiting the benefits of substantially larger datasets.

\begin{figure}[hbtp!]
    \centering
    \includegraphics[width=0.6\linewidth]{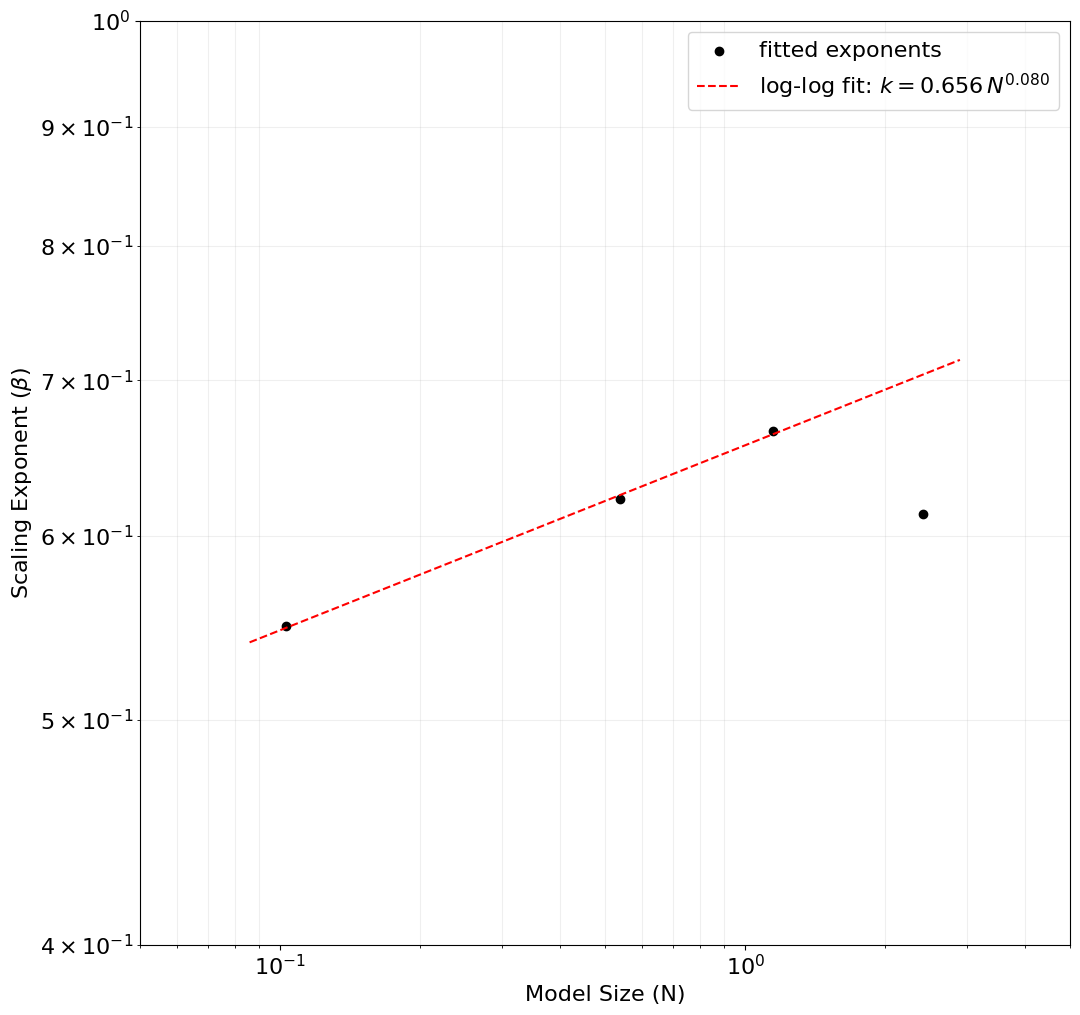}
    \caption{Exponent ($\beta$) scaling with model size (N) in millions.}
    \label{fig:beta-vs-N}
\end{figure}

Finally, we assess how the fitted scaling exponent varies with model size. Figure~\ref{fig:beta-vs-N} shows the scaling relationship. We observe an approximately linear correlation between $\beta$ and $N$ on a log-log scale when the model size is not excessively large. The Large model, however, deviates from this trend. The exact cause of this deviation requires further investigation, but one plausible explanation is that the model capacity exceeds what can be effectively optimized under the available computational budget. As shown in Table~\ref{tab:model-hpoptimized}, the Large model employs a learning rate that is an order of magnitude smaller than those of the smaller models and achieves a higher test \ac{MSE} than the Medium model, which contradicts typical expectations from model-size scaling \cite{kaplan2020scalinglawsneurallanguage, hoffmann2022trainingcomputeoptimallargelanguage}. We therefore interpret the selected hyper-parameters, although optimized using the same systematic procedure as for the other models, are compromises between training stability and performance under a fixed computational budget. It is likely that the Large model's capability exceeds what is permitted by the optimizer to drive convergence beyond local minimum under $C^*$, and as a result the model's representational capability is optimization-limited. Future work will extend the present data-scaling study to model trainings with higher computational budgets to better assess scaling behavior for larger models.

If the computational budget limitation is indeed the cause of the lower fitted $\beta$ for the Large model, we would expect that increasing the computational budget $C^*$ would lead to an increase in the estimation of $\beta$'s magnitude for the Large model, conditioned on proper adjustments of training hyper-parameters. Focusing on the overall trend in Figure~\ref{fig:beta-vs-N}, we conjecture that larger models tend to exhibit larger magnitudes of $\beta$, indicating faster convergence and more efficient utilization of training data, until the allocated computational budget becomes the limiting factor. Such behavior of larger models are more efficient to train \cite{rosenfeld2019constructivepredictiongeneralizationerror} under limited computational budget is supported by other studies in the language modeling domain \cite{pmlr-v202-geiping23a, izsak-etal-2021-train}, and this is due to the greater capacity of larger models to learn task-adaptive representations that extract more features from each data sample. This conjecture suggests that even for domain-specific surrogate models, such as \texttt{MF-VortexNet}, there are benefits of constructing larger and wider models: increased model capacity not only enhances field representation and generalization across flow and geometric configurations, but also improves data utilization efficiency during training. As a consequence, larger models may require fewer training samples to achieve a target accuracy, potentially reducing the cost of dataset construction.

However, larger models also demand higher computational budgets for training. In practice, particularly during the conceptual design phase, computational resources are often pre-allocated. The optimal allocation of computational budget between dataset construction and model training when using domain-specific surrogate models therefore depends on the relative costs of generating \ac{CFD} datasets and training surrogate models, and should be determined by jointly considering these costs together with the training-efficiency gains presented by model size scaling.

\section{Conclusion and Future Work}
\label{sec:conclusion}

This study addresses two gaps in data-driven surrogate modeling for aerodynamic design: the limited presence of open-source, multi-fidelity datasets for flow simulations with complex physics over relatively simple geometries, and the lack of quantification of the empirical scaling laws for field-predicting GNN-based surrogate models.

A primary contribution of this work is the release of an open-source, multi-fidelity aerodynamic dataset for a parametric family of double-delta wings. To support data-driven modeling of nonlinear, vortex-dominated flows, we generated 272 unique geometries and evaluated each configuration at $Ma=0.3$ and angles of attack ranging from $11^\circ$ to $19^\circ$ using both \ac{VLM} and RANS \ac{CFD} solvers. The \acf{DOE} is carefully designed to enable future expansion of the dataset and to support variance-based sensitivities analyses. This dataset is publicly available and intended to serve as a benchmark for multi-fidelity surrogate modeling research.

Using this dataset, we conduct a preliminary empirical scaling study of the \texttt{MF-VortexNet} surrogate model. The scaling analysis is performed on six datasets, containing between 8 and 256 double-delta wing designs distributed in a hierarchical manner through sequential refinement of a design space's sampling density. By selecting snapshots at specific free-stream conditions, we assembled six training datasets with sizes $D$ ranging from 40 to 1280 graphs. Four models, with sizes ranging from 0.1 to 2.4 million trainable parameters, are trained under a fixed computational budget of at most 8000 weight update steps, subject to early stopping criteria. Collectively, we estimate that the test performance $\epsilon'$, evaluated using the test \ac{MSE} on the holdout test set, scales with $D$ as $\epsilon' \propto D^{-0.6122}$. The observed scaling dependency is strong, indicating efficient utilization of the training data. Leveraging this scaling exponent, and benchmarking against the minimal test \ac{MSE} achieved using a larger training dataset of size $2048$, we estimate that the optimal geometric sampling distance for \texttt{MF-VortexNet} under the current computational budget is approximately $0.34$ in a six-dimensional design space, which corresponds to a sampling density of eight samples along each dimension using space-filling sampling techniques. The relatively sparse sampling requirement supports the design intent of \texttt{MF-VortexNet} as an efficient multi-fidelity surrogate model for domain-specific applications. 

By comparing the fitted data size scaling exponents across differently sized models trained under a fixed computational budget, we observe a log-log linear correlation between the model size and data sizing exponent, although the largest model trained in the current study does not follow this trend. We speculate that this anomaly in the largest model is caused by insufficient computational budget, resulting in a set of suboptimal hyper-parameters for training the model. We conjecture that, when sufficient computational resources are allocated, the magnitude of data scaling exponent increases with larger model size, thereby allowing more efficient utilization of the dataset for larger model. Following this conjecture, larger surrogate model could be trained with less training data, yielding lower cost for offline dataset generation, at the expense of increased model training computation requirements. Consequently, there exists a trade-off between the computational budget allocated to dataset generation and that allocated to model training. Further study is required to systematically vary the computational budget for surrogate model training and to evaluate additional model sizes, in order to obtain more statistically robust estimates of the power-law exponents and to better characterize this trade-off.

In the future, several extensions of the existing work can be made. Firstly, downstream applications, such as sensitivity assessments of surrogate model performance with respect to geometric design variables, can be performed leveraging the \ac{DOE} used for the dataset. In addition, as approximately $2.3\%$ of the snapshots in the dataset require higher-fidelity \ac{CFD} solver to achieve convergence, incorporating higher fidelity solutions into the dataset may further enrich its fidelity levels. The preliminary assessment of the data-size scaling relationship conducted in this study indicates that the scaling-law analysis is incomplete without systematically varying the computational budget along with data and model size scaling. In future work, we aim to jointly scale the computational budget, model size, and dataset size to identify optimal resource-allocation strategies.

\section{Acknowledgment}

This work was supported by NASA, Award 80NSSC19K1661, under the Commercial Supersonics Technology (CST) program, Supersonic Configurations at Low Speeds (SCALOS), with Sarah Langston as the NASA technical grant monitor. The authors extend their gratitude to Kuang-Ying Ting, Eli Livne, and Chester Nelson for their insightful discussions on the CST design space, and Aurelien Ghiglino for his discussion on the scaling experiments. Additionally, the authors would like to thank Stanford Research Computing Center for providing computational resources on the Sherlock cluster, as well as the Google Cloud Research Credit Program for granting access to GPU resources on the Google Cloud Platform.

\newpage
\bibliography{scalos}

\end{document}